\documentclass{article}

\usepackage{microtype}
\usepackage{graphicx}
\usepackage{booktabs} %
\usepackage{caption}
\usepackage{subcaption}
\usepackage{verbatim}

\usepackage{hyperref}

\newcommand{\theHalgorithm}{\arabic{algorithm}}

\usepackage[accepted]{icml2020}

\usepackage{amsmath,amsfonts,bm}

\def\eqref#1{equation~\ref{#1}}

\def\1{\bm{1}}

\def\vtheta{{\bm{\theta}}}

\def\vi{{\bm{i}}}

\def\vl{{\bm{l}}}
\def\vm{{\bm{m}}}

\def\vo{{\bm{o}}}

\def\vy{{\bm{y}}}

\DeclareMathAlphabet{\mathsfit}{\encodingdefault}{\sfdefault}{m}{sl}
\SetMathAlphabet{\mathsfit}{bold}{\encodingdefault}{\sfdefault}{bx}{n}

\begin{document}

\twocolumn[
\icmltitle{Entropy Minimization In Emergent Languages}

\begin{icmlauthorlist}
\icmlauthor{Eugene Kharitonov}{fair}
\icmlauthor{Rahma Chaabouni}{fair,inria}
\icmlauthor{Diane Bouchacourt}{fair}
\icmlauthor{Marco Baroni}{fair,icrea}
\end{icmlauthorlist}

\icmlaffiliation{fair}{Facebook AI Research, Paris, France}
\icmlaffiliation{icrea}{Catalan Institute for Research and Advanced Studies, Barcelona, Spain}
\icmlaffiliation{inria}{Cognitive Machine Learning (ENS - EHESS - PSL - CNRS - INRIA)}

\icmlcorrespondingauthor{Eugene Kharitonov}{kharitonov@fb.com}

\vskip 0.3in
]

\printAffiliationsAndNotice{}  %

\begin{abstract}
There is growing interest in studying the languages that emerge when neural agents are jointly trained to solve tasks requiring communication through a discrete channel.  We investigate here the information-theoretic complexity of such languages, focusing on the basic two-agent, one-exchange setup. We find that, under common training procedures, the emergent languages are subject to an entropy minimization pressure that has also been detected in human language, whereby the mutual information between the communicating agent's inputs and the messages is minimized, within the range afforded by the need for successful communication. That is, emergent languages are (nearly) as simple as the task they are developed for allow them to be. This pressure is amplified as we increase communication channel discreteness. Further, we observe that stronger discrete-channel-driven entropy minimization leads to representations with increased robustness to overfitting and adversarial attacks. We conclude by discussing the implications of our findings for the study of natural and artificial communication systems.
\end{abstract}

\section{Introduction}
\label{sIntroduction}

There has recently been much interest in the analysis of the
communication systems arising when %
deep network agents that interact to accomplish a goal are allowed to
exchange language-like discrete messages
~\citep{Lazaridou2016,Havrylov2017,Choi2018,Lazaridou2018,Li2019ease,Chaabouni2020}. Understanding
the emergent protocol is important if we want to eventually develop
agents capable of interacting with each other and with us through
language \citep{Mikolov2016,ChevalierBoisvert:etal:2019}. The pursuit
might also provide comparative evidence about how core properties of
human language have evolved
\citep{Kirby2002,Hurford:2014,Graesser:etal:2019}. While earlier
studies reported ways in which deep agent protocols radically depart
from human language
\citep{Kottur2017,Bouchacourt2018,Chaabouni:etal:2019,Lowe:etal:2019},
we show here that emergent communication shares an important property
of the latter, namely a %
tendency towards~\emph{entropy~minimization}.

Converging evidence shows that efficiency pressures are at work in
language and other biological communication systems
\citep{FerrerICancho:etal:2013,Gibson:etal:2019}. One particular
aspect of communicative efficiency, robustly observed
across many semantic domains, is the tendency to minimize lexicon entropy, to the extent allowed by the
counteracting need for accuracy
\citep{Zaslavsky:etal:2018,Zaslavsky:etal:2019}. For example, while
most languages distinguish grandmothers from grandfathers, few
have separate words for mother- and father-side grandmothers, as the
latter distinction makes communication only slightly more
accurate at the cost of an increase in lexicon complexity
\citep{Kemp:Regier:2012}. We show here, in two separate games designed
to precisely measure such property, 
that the protocol evolved by interacting deep agents is
subject to the same complexity minimization~pressure. 

Entropy minimization in natural language has been connected to the
Information Bottleneck principle~\citep{Tishby1999}. In turn, 
complexity reduction due to the Information Bottleneck provides a beneficial regularization effect on learned representations~\citep{Fischer2019,Alemi2016,Achille2018,Achille2018b}. It is difficult to experimentally
verify the presence of such effect in human language, but we can look
for it in our computational simulations. We confirm that, when
relaxing channel discreteness, the entropy minimization property
no longer holds, and the system becomes less robust against
overfitting and adversarial noise. This in turn raises intriguing questions about the origin of discreteness in human language, that we return to in the conclusion.

\section{General framework}

We establish our results in the context of signaling games~\citep{Lewis1969}, as
introduced to the current language emergence literature by
\citet{Lazaridou2016} and adopted in several later
studies~\citep{Havrylov2017,Bouchacourt2018,Lazaridou2018}. There are
two agents, Sender and Receiver, provided with individual inputs at
the beginning of each episode. Sender sends a single message to
Receiver, and Receiver has to perform an action based on its own input and
the received message. Importantly, there is no direct supervision on
the message protocol. We consider agents that are deterministic
functions of their inputs (after training).

As an example, consider the task of communicating a $n$-bit number, sampled uniformly at random from $0$...$2^{n}-1$. The full number is shown to Sender, and its $k$ ($0 \le k \le n$) least-significant bits are also revealed to Receiver. Receiver has to output the full number, based on the message from Sender and its own input.  Would Sender transmit the entire number through its message? In this case, the protocol
would be ``complex,'' encoding $n$ bits. Alternatively,
Sender could only encode the bits that Receiver does not know, and let Receiver fill in the rest by itself. This emergent protocol would be ``simple,''
encoding only strictly necessary information. We find experimentally that, once the agents are successfully trained to jointly solve the task, the emergent protocol \textit{minimizes the entropy of the messages} or, equivalently in our setup, \emph{the mutual information between Sender's input and messages}. In other words, the agents consistently
approximate the simplest successful protocol (in the current example,
the one transmitting $\approx n - k$ bits).

We can connect the entropies of Sender and Receiver inputs $\vi_s$ and $\vi_r$, messages $\vm$, Receiver's output (the chosen action) $\vo$, and ground-truth outputs $\vl$ by standard inequalities~\citep{Cover2012}.\footnote{We also use the fact that that  $H(x) \ge H(g(x))$ for any discrete r.v.\ $x$ and function $g$.} Denoting Sender's computation as a function $S: S(\vi_s) = \vm$, and Receiver as function $R: R(\vm, \vi_r) = \vo$, we obtain:
\begin{multline}
H(\vi_s) \ge H(S(\vi_s)) =  H(\vm) \ge H(\vm | \vi_r) \ge \\ \ge H(R(\vm, \vi_r) | \vi_r) = H(\vo | \vi_r)  \approx H(\vl | \vi_r),
\end{multline}
where the last relation stems from the fact that after successful training $\vo \approx \vl$. %
Note that, since agents are deterministic after training, $H(\vm) = I(\vi_s; \vm)$. We can then use these quantities interchangeably.

Our empirical measurements indicate that the entropy of the  messages $\vm$ in the emergent protocol tends to approach the lower bound: $H(\vm) \rightarrow H(\vl | \vi_r)$, even if the upper bound $H(\vi_s)$ is far. %
that Receiver needs is reduced   without changing other parameters, the emergent protocol becomes simpler (lower entropy). In other words, the  emergent protocol adapts to minimize the information that passes through~it.

Code for our experiments is publicly available at %
\href{https://github.com/facebookresearch/EGG/}{github.com/facebookresearch/EGG/} as a part of the EGG framework~\cite{Kharitonov2019Egg}.
\section{Methodology}
\label{sMethodology}

\subsection{Games}
\label{sTasks}

We study two signaling games. In Guess Number, the agents are trained to recover an integer-representing vector with uniform Bernoulli-distributed components. This simple setup gives us full control over the amount of information needed to solve the task. The second game, Image Classification, employs more naturalistic data, as the agents are jointly trained to classify pairs of MNIST digits \citep{Lecun1998}.

\textbf{Guess Number} We draw an 8-bit integer $0 \le z \le 255$
uniformly at random, by sampling its 8 bits independently from the
uniform Bernoulli distribution. All bits are revealed to Sender as an
8-dimensional binary vector $\vi_s$. The last $k$ bits are revealed to
Receiver ($0 \le k \le 8$) as its input $\vi_r$. Sender outputs a
single-symbol message $\vm$ to Receiver. In turn, Receiver outputs a
vector $\vo$ that recovers all the bits of $z$ and should be equal to
$\vi_s$.

In this game, Sender has a linear layer that maps the input vector $\vi_s$ to a hidden representation of size 10, followed by a leaky ReLU activation. Next is a linear layer followed by a softmax over the vocabulary. Receiver linearly maps both its input $\vi_r$ and the message to 10-dimensional vectors, concatenates them, applies a fully connected layer with output size 20, followed by a leaky ReLU. Finally, another linear layer and a sigmoid nonlinearity are applied. When training with REINFORCE and the Stochastic Computation graph approach (see Sec.~\ref{ssTraining}), we increase the hidden layer sizes threefold, as this leads to a more robust convergence.

\textbf{Image Classification}
In this game, the agents are jointly trained to classify 28x56 images of two MNIST digits, stacked side-by-side (more details in Supplementary). Unlike Guess Number, Receiver has no side input. Instead, we control the informational complexity of Receiver's task by controlling the size of its output space, i.e., the number of labels we assign to the images. To do so, we group all two-digit sequences $00..99$ into  $N_l \in \{2, 4, 10, 20, 25, 50, 100\}$ equally-sized classes.

In Sender, input images are embedded by a LeNet-1 instance~\citep{Lecun1990} into 400-dimensional vectors. These embedded vectors are passed to a fully connected layer, followed by a softmax selecting a vocabulary symbol. Receiver embeds the received messages into 400-dimensional vectors, passed to a fully connected layer with a softmax activation returning the class probabilities.

We report hyperparameter grids in Supplementary. In the following experiments, we fix vocabulary to 1024 symbols (experiments
with other vocabulary sizes, multi-symbol messages, and larger architectures are reported in Supplementary). No parts of the agents are pre-trained or shared.  The loss being optimized depends on the chosen gradient estimation method (see
Sec.~\ref{ssTraining}). We denote it $\mathcal{L}(\vo, \vl)$, and
it is a function of Receiver's output $\vo$ and the ground-truth
output $\vl$.  When training in Guess Number with REINFORCE, we use a 0/1 loss: the
agents get zero loss only when all bits of $z$ are correctly recovered. When training with  Gumbel-Softmax relaxation or the Stochastic Computation Graph approach, we use binary
cross-entropy (Guess Number) and negative log-likelihood
(Image Classification).

\subsection{Training with discrete channel}
\label{ssTraining}
Training to communicate with discrete messages is non-trivial, as
we cannot back-propagate through the messages. Current
language emergence work mostly uses Gumbel-Softmax
relaxation \citep[e.g.,][]{Havrylov2017}
or REINFORCE \citep[e.g.,][]{Lazaridou2016} to get
gradient estimates.  We also explore the Stochastic Computation Graph
optimization approach. We plug the obtained gradient estimates into Adam~\citep{Kingma2014}.

\textbf{Gumbel-Softmax relaxation} Samples from the Gumbel-Softmax distribution (a) are reperameterizable, hence allow gradient-based training, and (b) approximate samples from the corresponding Categorical distribution~\citep{Maddison2016,Jang2016}. To get a sample that approximates an $n$-dimensional Categorical distribution with probabilities $p_i$, we draw $n$ i.i.d.\ samples $g_i$ from Gumbel(0,1) and use them to calculate a vector $\vy$ with components:
\begin{equation}
\label{eq:gs-reparam}
y_i = \frac{exp\left[(g_i + \log ~ p_i) / \tau \right]}{\sum_j exp\left[(g_j + \log ~ p_j) / \tau\right]},
\end{equation}
where $\tau$ is the temperature hyperparameter. As $\tau$ tends to $0$, the samples $\vy$ get closer to one-hot samples; as $\tau \rightarrow +\infty$, the components $y_i$ become uniform. During training, we use these relaxed samples as messages from Sender, making the entire Sender/Receiver setup differentiable.

\textbf{REINFORCE} by \citet{Williams1992} is a standard reinforcement learning algorithm. In our setup, it estimates the gradient of the expectation of the loss $\mathcal{L}(\vo, \vl)$ w.r.t.\ the parameter vector $\vtheta$ as follows:
\begin{equation}
    \label{eq:reinforce}
    \mathbb{E}_{\vi_s, \vi_r} \mathbb{E}_{\vm \sim S(\vi_s), \vo \sim R(\vm, \vi_r)} \left[ (\mathcal{L}(\vo; \vl) - b) \nabla_{\vtheta} \log P_{\vtheta}(\vm, \vo) \right]
\end{equation}
The expectations are estimated by sampling $\vm$ from Sender and, after that, sampling $\vo$ from Receiver. We use the running mean baseline $b$~\citep{Greensmith2004,Williams1992} as a control variate. %
We adopt the common trick to add an entropy regularization term~\citep{Williams1991,Mnih2016} that favors higher entropy. We impose entropy regularization on the outputs of the agents with coefficients $\lambda_s$ (Sender) and $\lambda_r$ (Receiver).

\textbf{Stochastic Computation Graph (SCG)} In our setup, the gradient estimate approach of~\citet{Schulman2015} reduces to computing the gradient of the surrogate function:
\begin{equation}
    \label{eq:surrogate}
    \mathbb{E}_{\vi_s, \vi_r} \mathbb{E}_{\vm \sim S(\vi_s)} \left[\mathcal{L}(\vo; \vl) + sg \left( \mathcal{L}(\vo; \vl) - b \right) \log P_{\vtheta}(\vm) \right],
\end{equation}
where $sg$ denotes \textit{stop-gradient} operation.
We do not sample Receiver actions: Its parameter gradients are obtained with standard backpropagation (first term in Eq.~\ref{eq:surrogate}). Sender's messages are sampled, and its gradient is calculated akin to REINFORCE (second term in Eq.~\ref{eq:surrogate}). Again, we apply entropy-favoring regularization on Sender's output (with coefficient $\lambda_s$) and use the mean baseline.

\textbf{Role of entropy regularization} As we mentioned above, when training with REINFORCE and SCG, we include a (standard) entropy regularization term in the loss which explicitly \textit{maximizes} entropy of Sender's output. Clearly, this term is at odds with the entropy \textit{minimization} effect we observe. In our experiments, we found that high values of $\lambda_s$ (the parameter controlling Sender's entropy regularization) prevent communication success; on the other hand, a small non-zero $\lambda_s$ is crucial for successful training. In Sec.~\ref{sExperiments} we investigate the effect of $\lambda_s$ on entropy minimization.\footnote{The parameter $\lambda_r$, that controls Receiver's entropy regularization, does not influence the observed effect.}

\subsection{Experimental protocol}
In Guess Number, we use all $2^8$  possible inputs for training, early stopping and analysis. In Image Classification, we train on random image pairs from the MNIST training data, and use image pairs from the MNIST held-out set for validation. 
We select the runs that achieved a high level of performance (training accuracy above 0.99 for Guess Number and validation accuracy above 0.98 for Image Classification), thus studying typical agent behavior \textit{provided they succeeded at the game}.

At test time, we select the Sender's message symbol greedily, hence the messages are discrete and Sender represents a (deterministic) function $S$ of its input $\vi_s$, $\vm = S(\vi)$. 
Calculating the entropy $H(\vm)$ of the distribution of discrete messages $\vm$ is straightforward. In Guess Number, we enumerate all 256 possible values of $z$ as inputs, obtain messages and calculate entropy $H(\vm)$. For Image Classification, we sample image pairs from the held-out set.

The upper bound on $H(\vm)$ is as follow: $H_{max} = 8$
bits (bounded by $H(\vi_s)$) in Guess Number, and $H_{max} = 10$ bits (bounded by vocabulary size) in
Image Classification. Its lower bound is equal to $H_{min} = H(\vl | \vi_r)= 8 - k$ bits for Guess number. In Image Classification, communication can only succeed if $H(\vm)$ is not less than $H(\vl)$, i.e.,
$H_{min} = H(\vl) = \log_2 N_{l}$, with $N_{l}$ the number of
equally-sized classes we split the images into.

\label{ssMinimization}
\begin{figure*}[t!]
  \centering
  \begin{subfigure}{0.5\linewidth}
  \centering
  \includegraphics[width=0.8\linewidth]{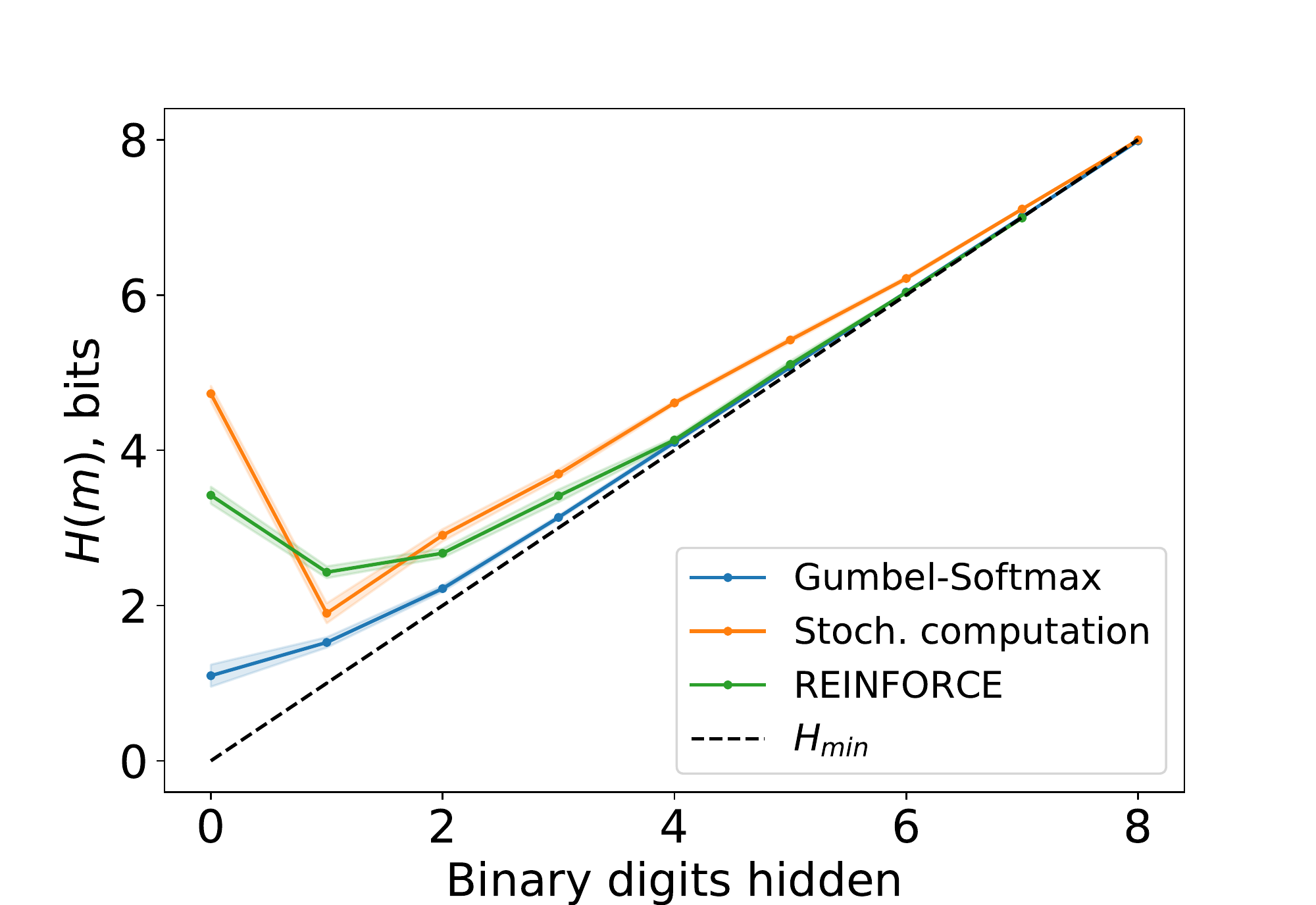}
\caption{All three training approaches.\newline}
    \label{fig:number:all}
  \end{subfigure}%
\begin{subfigure}{0.5\linewidth}
\centering
  \includegraphics[width=0.8\linewidth]{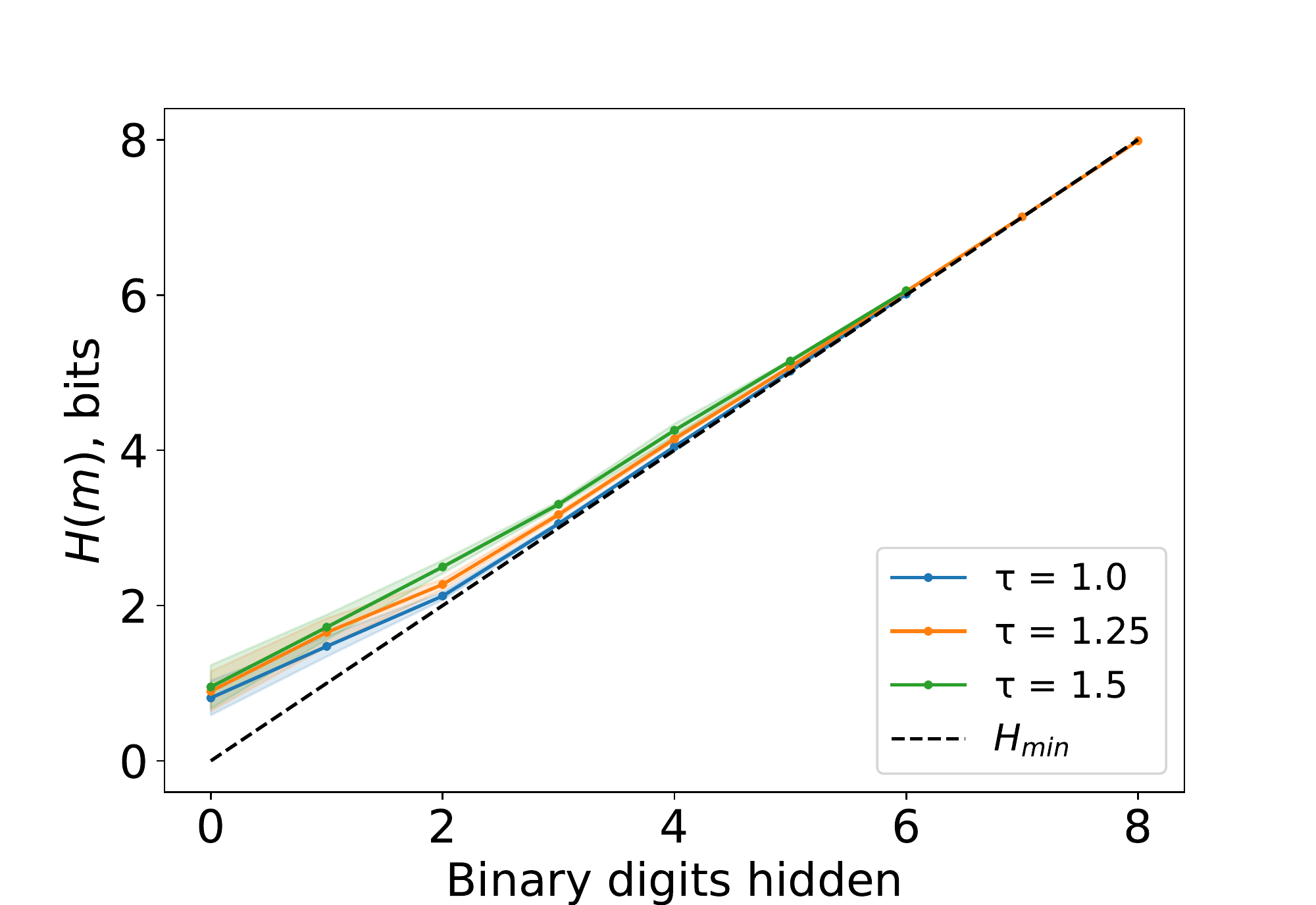}
\caption{Training with Gumbel-Softmax relaxation.}
  \label{fig:number:gs}
 \end{subfigure}

\begin{subfigure}{0.5\linewidth}
\centering
  \includegraphics[width=0.8\linewidth]{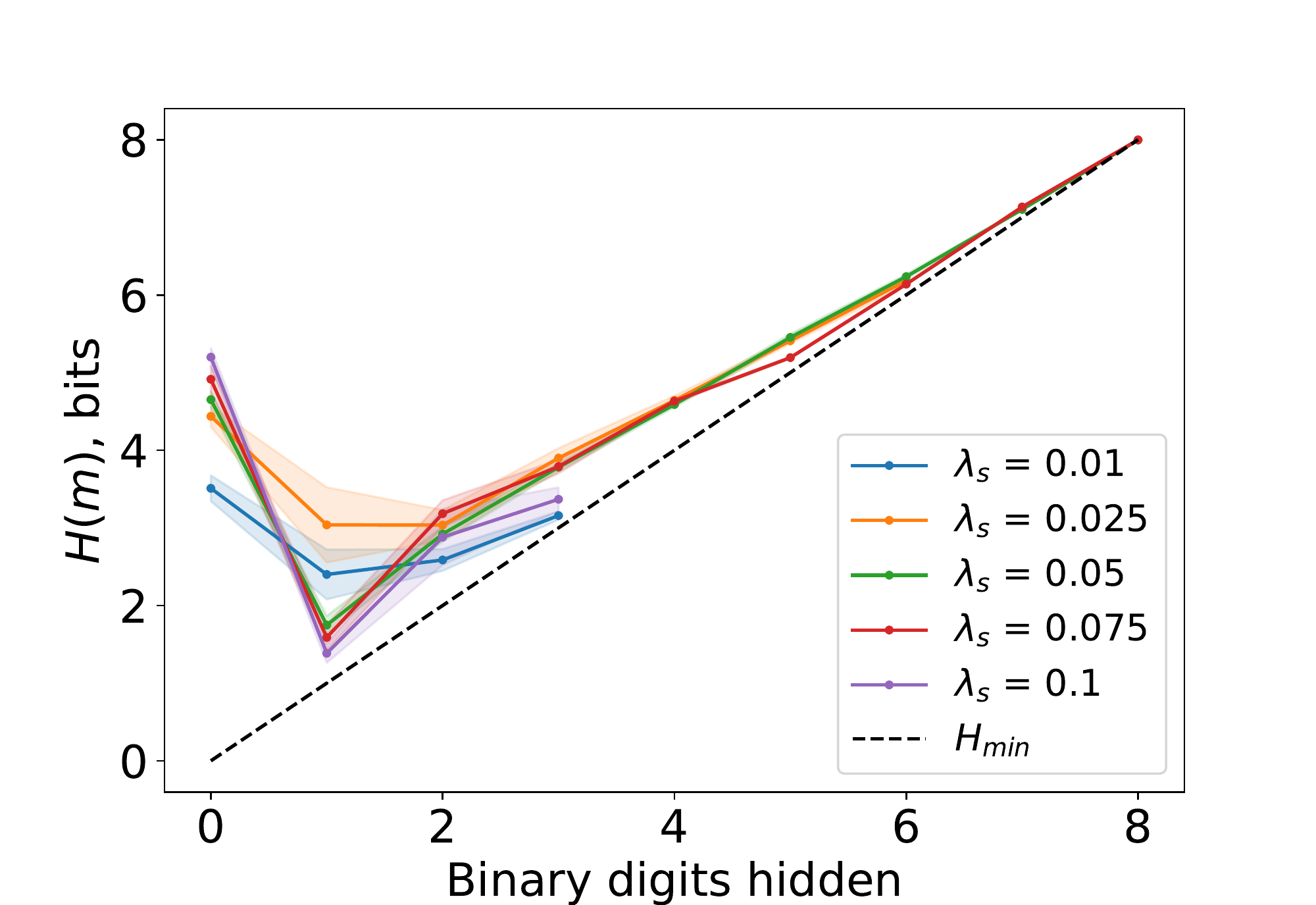}
  \caption{Training with Stochastic Computation Graph.}
  \label{fig:number:stoch}
 \end{subfigure}%
  \begin{subfigure}{0.5\linewidth}
\centering
  \includegraphics[width=0.8\linewidth]{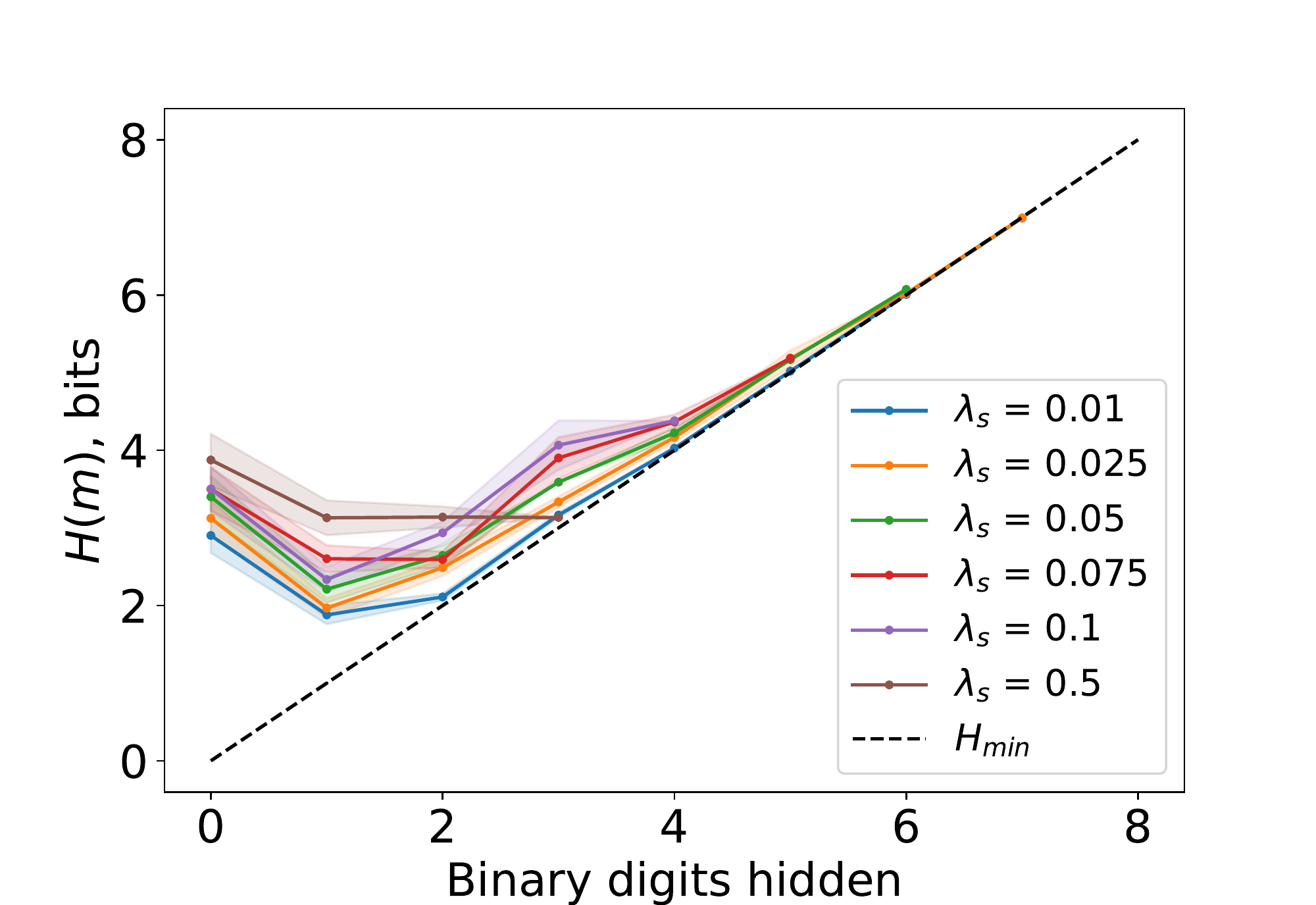}
  \caption{Training with REINFORCE.\newline}
  \label{fig:number:rf}
 \end{subfigure}
\caption{Guess Number: entropy of the messages $\vm$. Shaded regions represent one standard error of the mean (SEM).}
\label{fig:guess_number}
\end{figure*}

\begin{figure*}
\centering
\begin{subfigure}{0.5\linewidth}
\centering
\vspace{0.3cm}
  \includegraphics[width=0.8\linewidth]{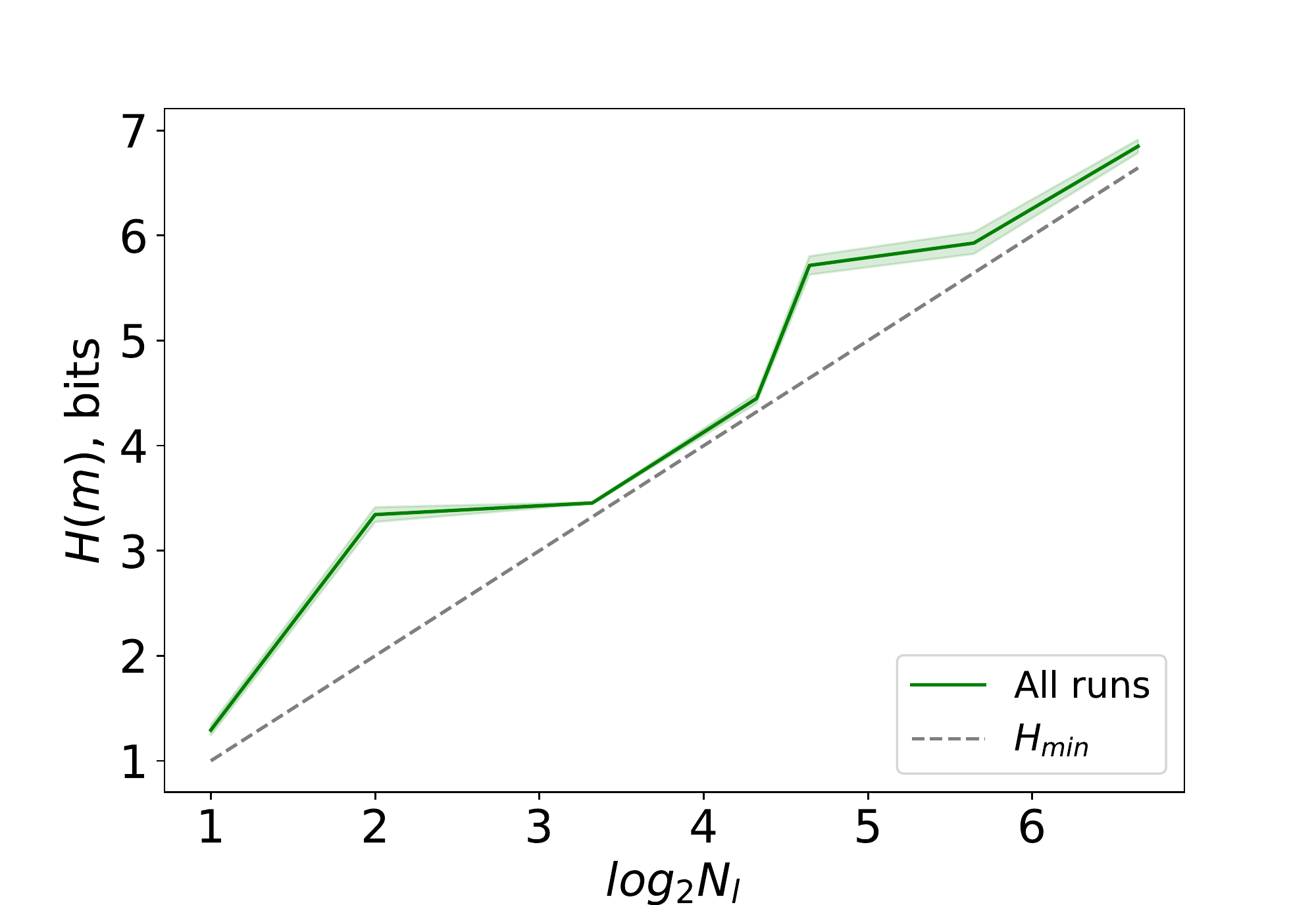}
  \caption{Successful runs pooled together.\newline}
  \label{fig:mnist:a}
 \end{subfigure}%
  \begin{subfigure}{0.5\linewidth}
\centering
  \includegraphics[width=0.8\linewidth]{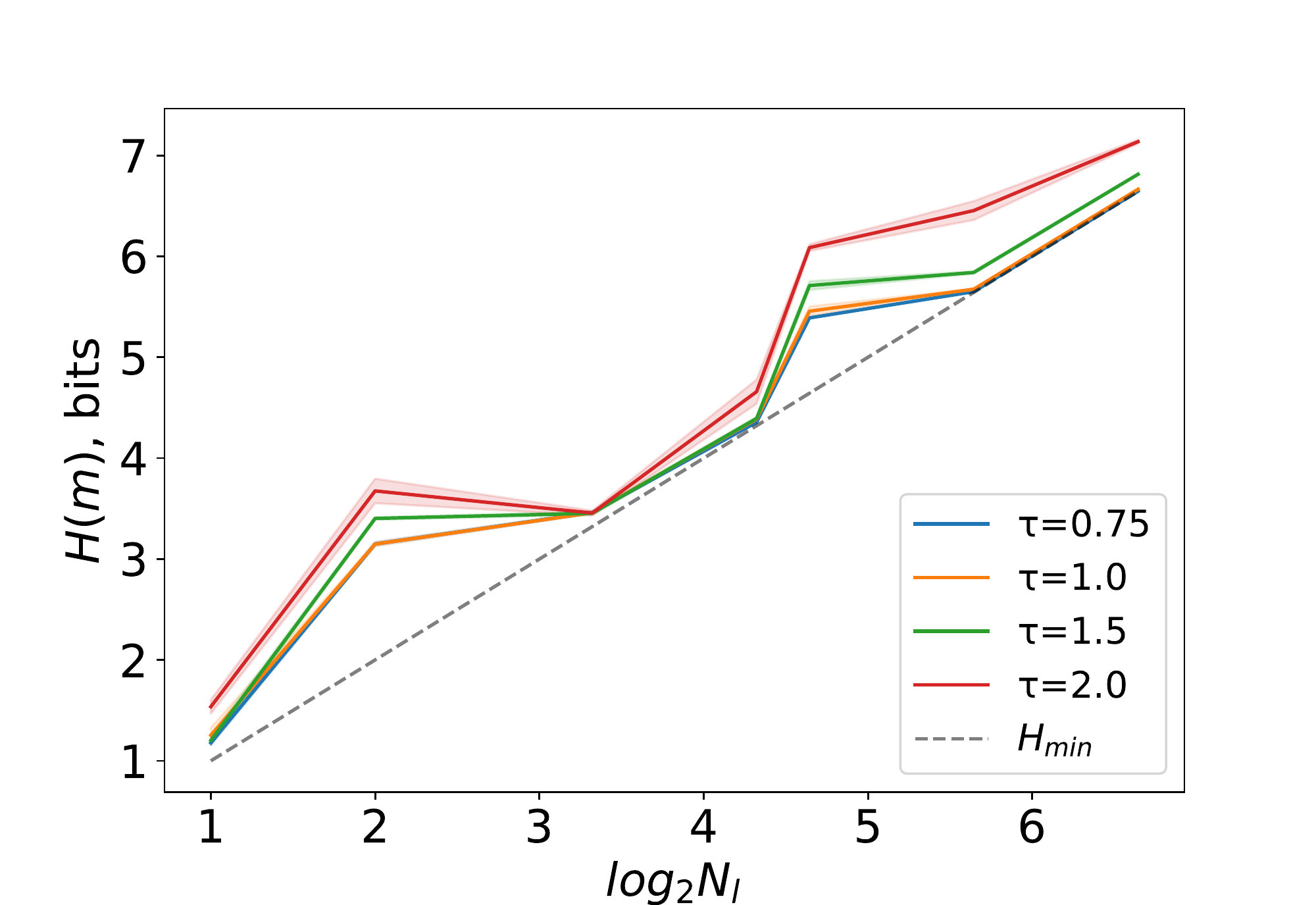}
  \caption{Successful runs grouped by temperature.}
  \label{fig:mnist:b}
 \end{subfigure}
\caption{Image Classification: entropy of the messages $\vm$ in function of log number of target classes, $N_l$. Shaded regions mark SEM.
}
  \label{fig:image_game}
\end{figure*}

\section{Experiments}
\label{sExperiments}

\subsection{Entropy minimization}
\textbf{Guess Number} In Figure~\ref{fig:guess_number}, the horizontal
axes span the number of bits of $z$ that Receiver
lacks, $8 - k$. The vertical axis reports the information content of the
protocol, measured by  messages entropy $H(\vm)$.
Each integer on the horizontal axis corresponds to a game configuration,
and for each such configuration we aggregate multiple (successful)
runs with different hyperparameters and random seeds. $H_{min}$ indicates the minimal amount of bits Sender has to send in a particular configuration for the task to be solvable. The upper bound (not shown) is $H_{max}=8$ bits. Across hyperparameters and random seeds, trainings with Gumbel-Softmax and SCG have success rate above 50\%. With REINFORCE success rate is approximately 20\%. %

Consider first the configurations where Receiver's input is insufficient to answer correctly (at least one binary digit hidden, $k \le 7$).  From Figure~\ref{fig:number:all}, we observe that the transmitted information is strictly monotonically  increasing with the number of binary digits hidden from Receiver. Thus, even if Sender sees the very same input in all configurations, a more nuanced protocol is only developed when it is necessary. Moreover, the entropy $H(\vm)$ (equivalently: the transmitted information) stays close to the lower bound. This entropy minimization property holds for all the considered training approaches across all configurations.

Consider next the configuration where Receiver is getting the whole integer $z$ as its input ($k=8$, the leftmost configuration in Figure~\ref{fig:guess_number}, corresponding to 0 on x axis). Based on the observations above, one would expect that the protocol would approach zero entropy in this case (as no information needs to be transmitted). However, the measurements indicate that the protocol is encoding considerably more information. It turns out that this information is entirely ignored by Receiver. To demonstrate this, we fed all possible distinct inputs to Sender, retrieved the corresponding messages, and \textit{shuffled} them to destroy any information about the inputs they might carry. The shuffled messages were then passed to Receiver alongside its own (un-shuffled) inputs. The overall performance was not affected by this manipulation, confirming the hypothesis that Receiver ignores the messages. We conclude that in this case there is no entropy minimization pressure on Sender simply because there is no communication. The full experiment is in~Supplementary.

We further consider the effect of various hyperparameters. In
Figure~\ref{fig:number:gs}, we split the results obtained with
Gumbel-Softmax by relaxation temperature. As discussed in
Sec.~\ref{ssTraining}, lower temperatures more closely approximate
discrete communication, hence providing a convenient control of the
level of discreteness imposed during training (recall that at test
time we enforce full discreteness by selecting the symbol greedily).  The figure shows that lower temperatures consistently
lead to lower $H( \vm)$.  This implies that, as we
increase the ``level of discreteness'' at training, we get
 stronger entropy minimization pressure.

In Figures~\ref{fig:number:stoch}~\&~\ref{fig:number:rf}, we report $H(\vm)$ when training with Stochastic Graph Optimization and REINFORCE across degrees of entropy regularization. We report curves corresponding to $\lambda_s$ values which converged in more than three configurations. With REINFORCE, we see a weak tendency for a higher $\lambda_s$ to trigger a higher entropy in the protocol. However, message entropy stays generally close to the lower bound even in presence of strong exploration, which favors higher entropy in  Sender's output distribution.

\textbf{Image Classification} As the models are more
complex, we only had consistent success when training with
Gumbel-Softmax (success rate is approximately 80\%).
In Figure~\ref{fig:mnist:a} we aggregate all successful runs. The
information encoded by the protocol grows as Receiver's output
requires more information. However, in all configurations, the
transmitted information stays well below the 10-bit upper bound and
tends to be close to $H_{min}$.  A natural interpretation is that
Sender prefers to take charge of image classification and directly pass
information about the output label, rather than sending along a
presumably more information-heavy description of the
input. In Figure~\ref{fig:mnist:b}, we split the runs by temperature. Again,
we see that lower temperatures consistently lead to stronger entropy
minimization pressures.

Summarizing, \textit{when communicating through a discrete channel, there is consistent pressure for the emergent protocol to encode as little information as necessary}. This holds across games,
training methods and hyperparameters. When training with Gumbel-Softmax,  temperature controls the strength of this pressure, confirming the relation between entropy minimization and discreteness.

\begin{figure*}
\centering
\begin{subfigure}{0.5\linewidth}
\centering
  \includegraphics[width=0.8\linewidth]{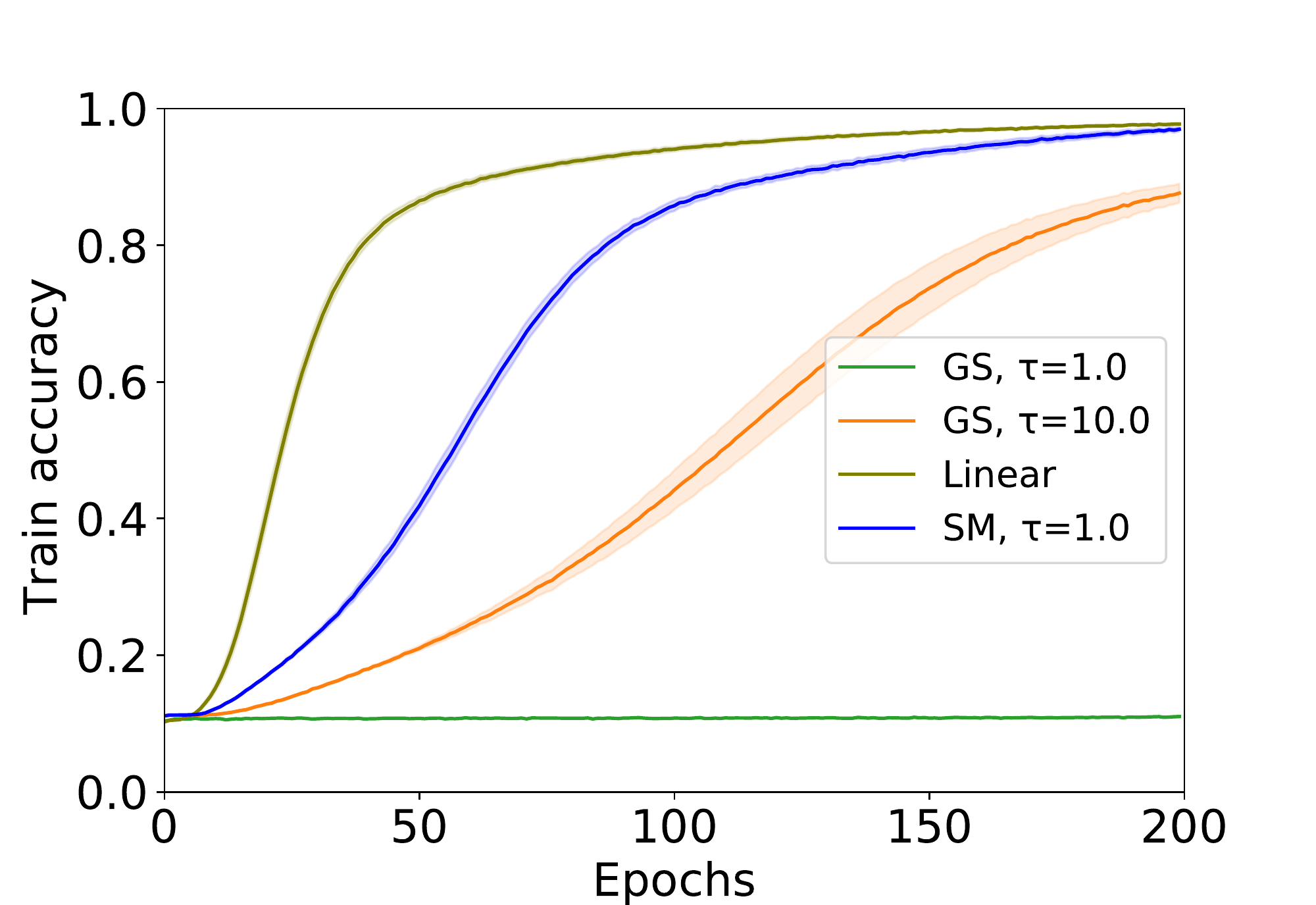}
\caption{All train labels are shuffled.}
  \label{fig:broken_mnist:c}
 \end{subfigure}%
\begin{subfigure}{0.5\linewidth}
\centering
  \includegraphics[width=0.8\linewidth]{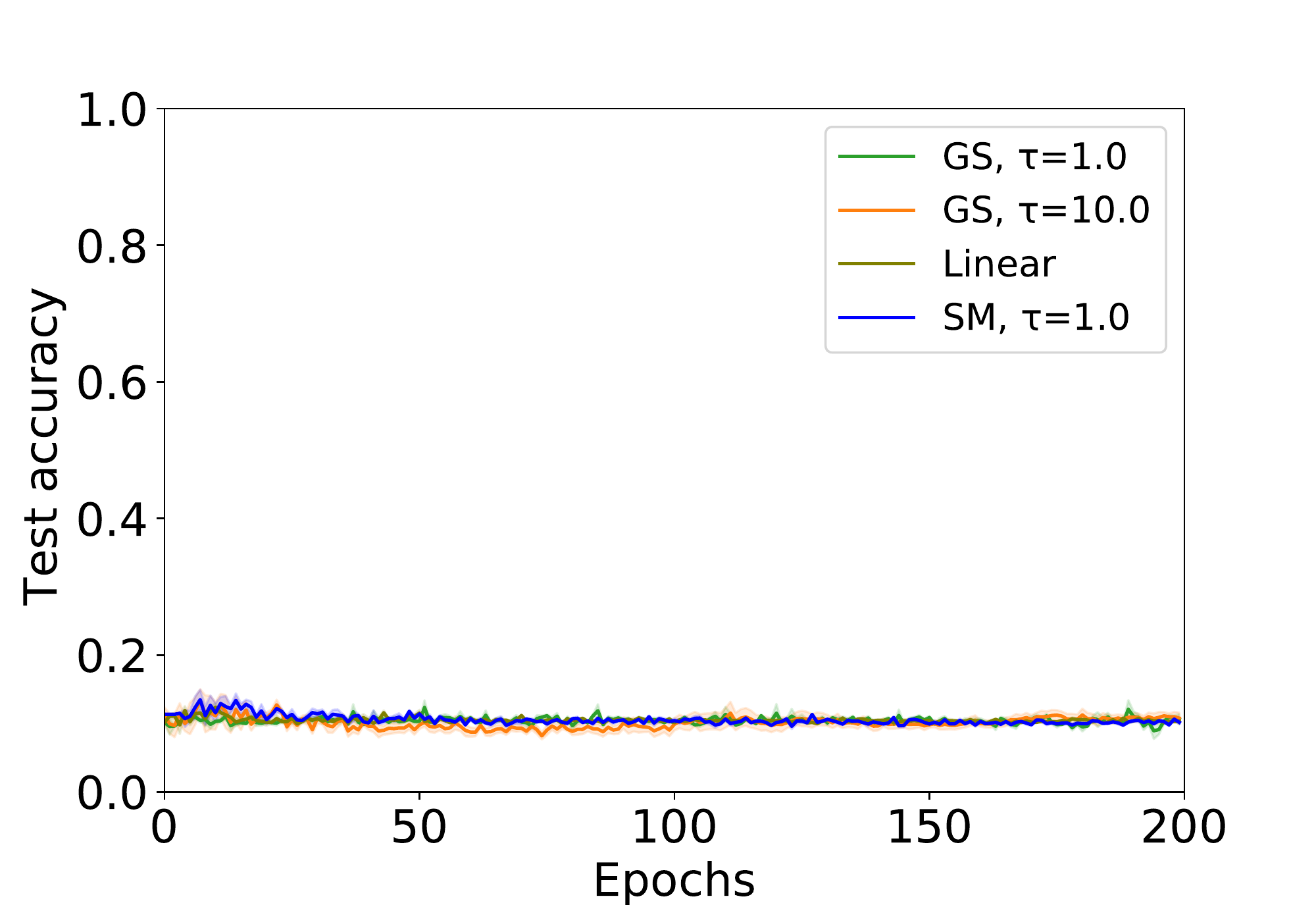}
\caption{All train labels are shuffled.}
  \label{fig:broken_mnist:d}
 \end{subfigure}

 \begin{subfigure}{0.5\linewidth}
\centering
  \includegraphics[width=0.8\linewidth]{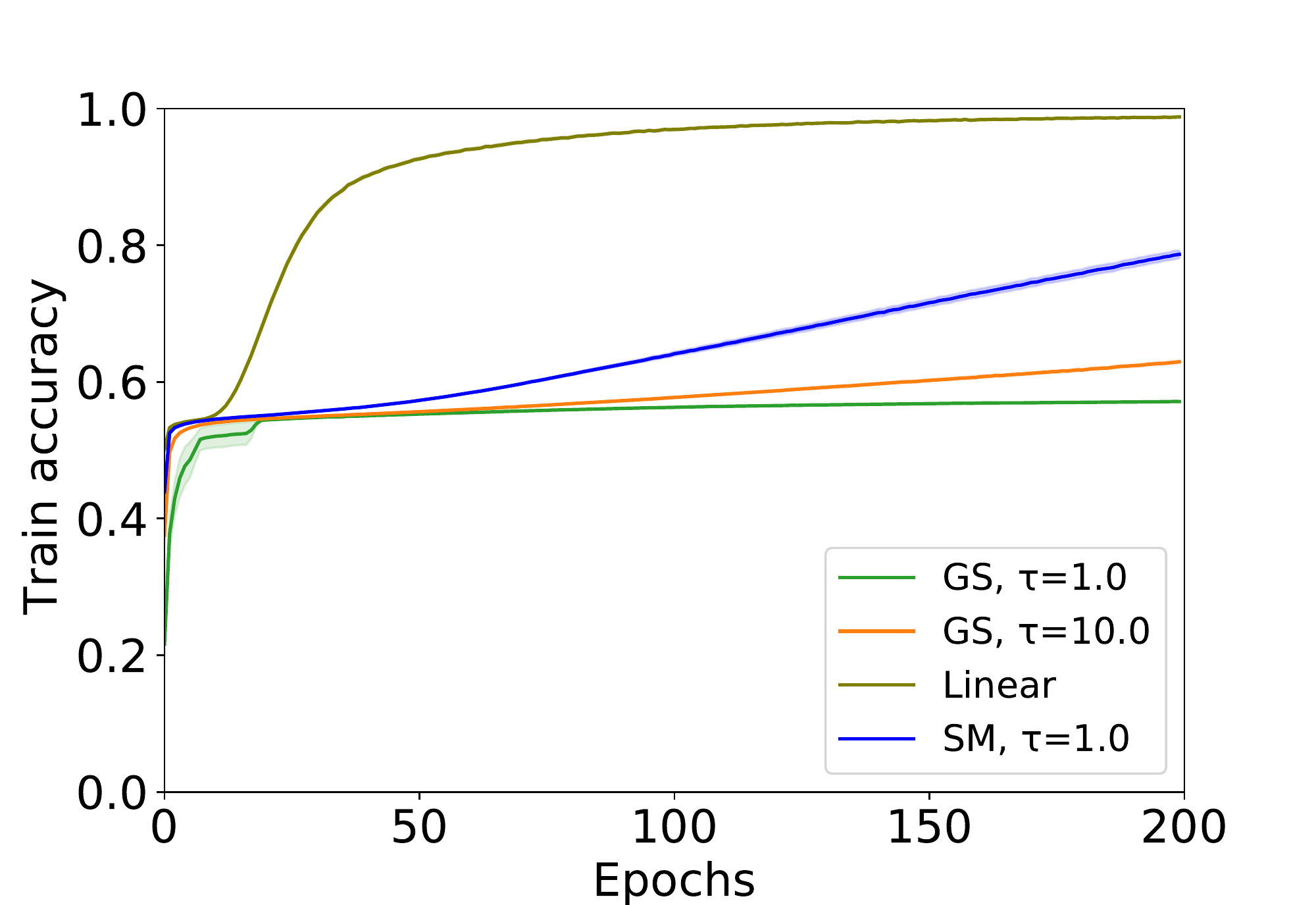}
\caption{Half of train labels are shuffled.}
  \label{fig:broken_mnist:e}
 \end{subfigure}%
\begin{subfigure}{0.5\linewidth}
\centering
  \includegraphics[width=0.8\linewidth]{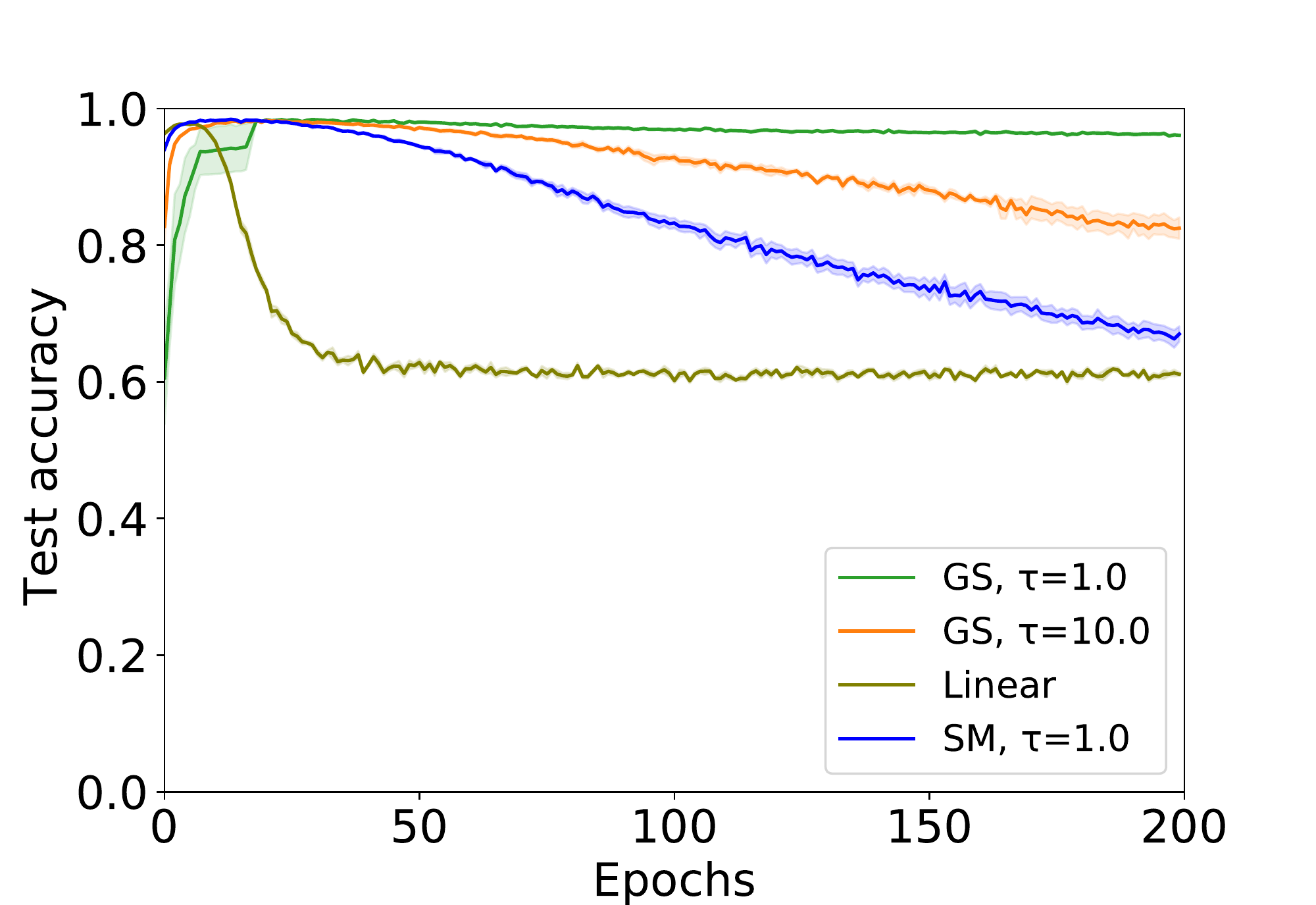}
\caption{Half of train labels are shuffled.}
  \label{fig:broken_mnist:f}
\end{subfigure}

\begin{subfigure}{0.5\linewidth}
\centering
  \includegraphics[width=0.8\linewidth]{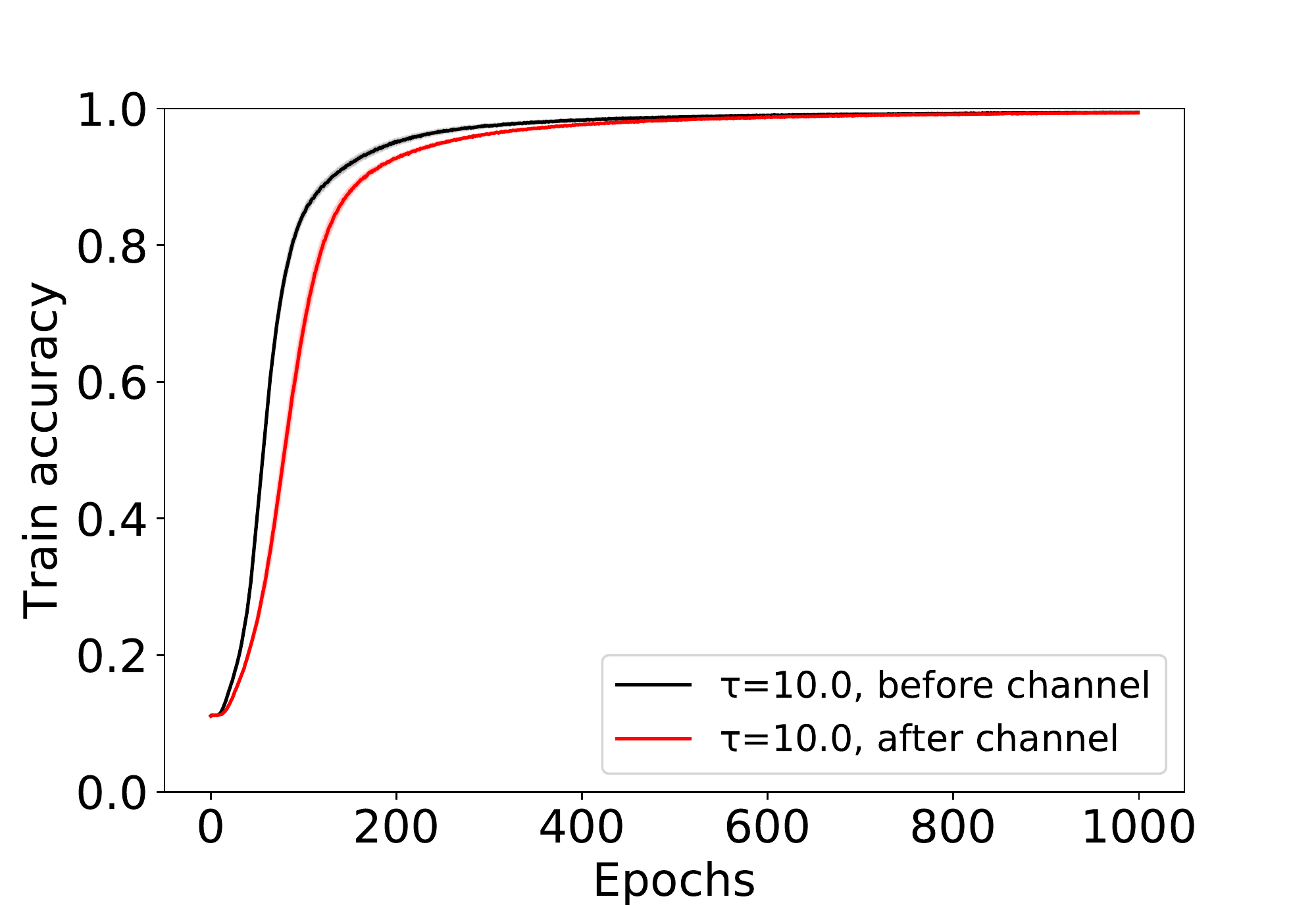}
\caption{All labels shuffled; Additional layer before channel vs.\ \newline after channel}
  \label{fig:broken_mnist:g}
 \end{subfigure}%
\begin{subfigure}{0.5\linewidth}
\centering
  \includegraphics[width=0.8\linewidth]{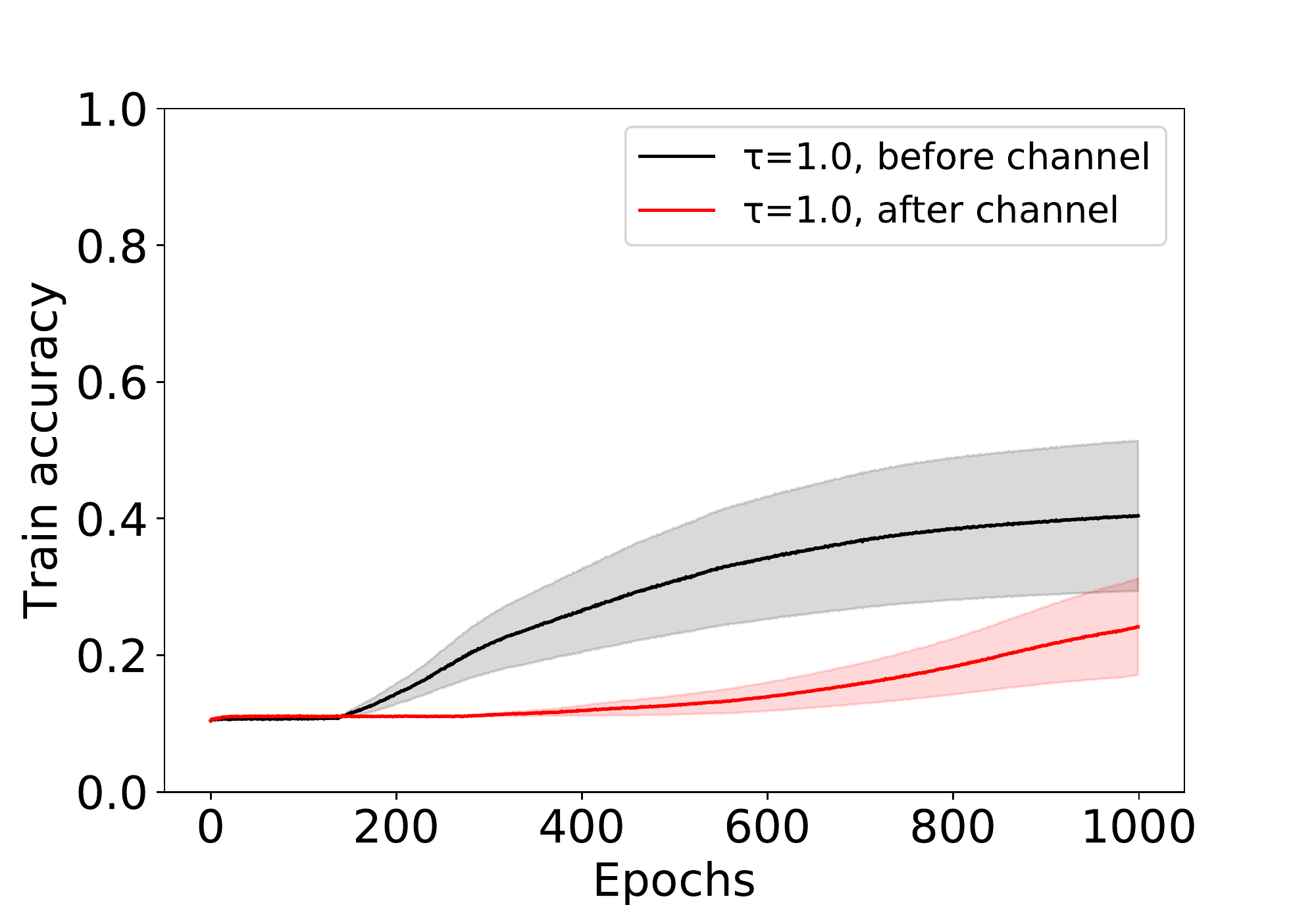}
\caption{All labels shuffled; Additional layer before channel vs.\ after channel}
  \label{fig:broken_mnist:h}
\end{subfigure}

\caption{Learning in presence of random labels. \textit{GS} (\textit{SM}) denotes models trained with Gumbel-Softmax  (Softmax) channel. \textit{Linear} are models with the channel removed.\vspace{-3mm}}
\label{fig:broken_mnist}
\end{figure*}

\subsection{Evolution of message entropy during training}
To gain further insights into the minimization trend, we studied the evolution of message entropy during training. We observed that the initial entropy of Sender can be both higher and lower than the minimum entropy $H_{min}$ required for solving the task. Further, we measured how the entropy of the messages changes after each training epoch by applying the same procedure as above, i.e., feeding the entire dataset to Sender and selecting the message symbol greedily. When message entropy starts higher than $H_{min}$, it falls close to it during the training. Similarly, when it starts lower than $H_{min}$, it increases during training. This experiment is reported in Supplementary. Thus, information minimization is not simply due to the difficulty of discovering a higher-entropy protocol during learning, but also due to the complexity of maintaining mutual coordination between the agents.

\subsection{Representation discreteness and robustness}
\label{ssRobustness}
The entropy minimization effect indicates that 
a discrete representation will only store as much information as necessary to solve the task. This emergent behavior resembles the Information Bottleneck principle \citep{Tishby1999,Achille2018}. The fact that lower training-time temperatures in Gumbel-Softmax optimization correlate with both higher discreteness and a tighter bottleneck (see Sec.~\ref{ssMinimization}) makes us further conjecture that discreteness is causally connected to the emergent bottleneck. The Information Bottleneck principle has also been claimed to govern
entropy minimization in natural language \citep{Zaslavsky:etal:2018,Zaslavsky:etal:2019}. Bottleneck effects in neural agents and 
 natural language might be due to the same cause, namely communication discreteness.

\begin{figure*}
\centering
\begin{subfigure}{0.5\linewidth}
\centering
  \includegraphics[width=0.8\linewidth]{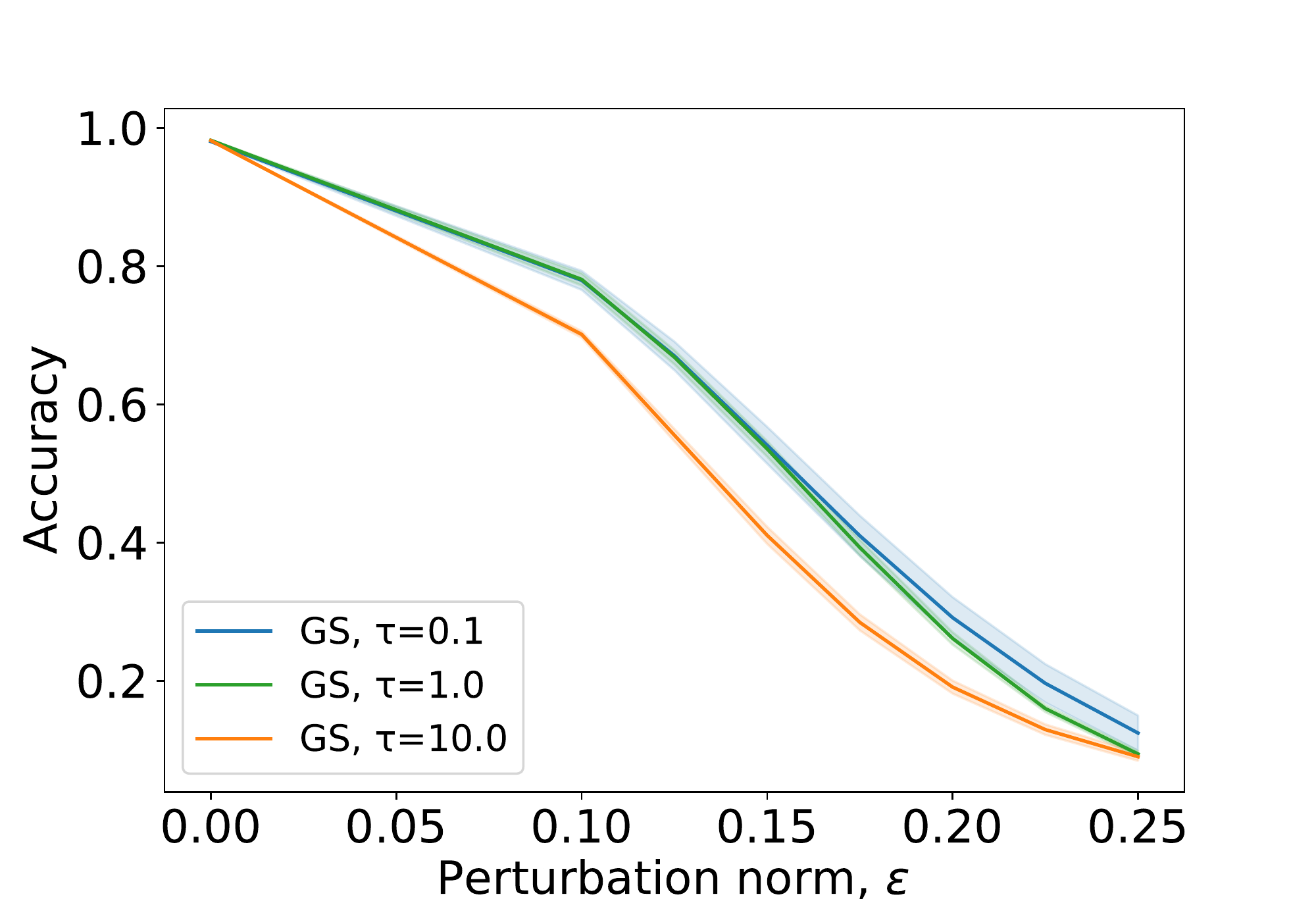}
\caption{Robustness vs.\ temp.\ $\tau$.}
  \label{fig:adversarial:a}
 \end{subfigure}%
\begin{subfigure}{0.5\linewidth}
\centering
  \includegraphics[width=0.8\linewidth]{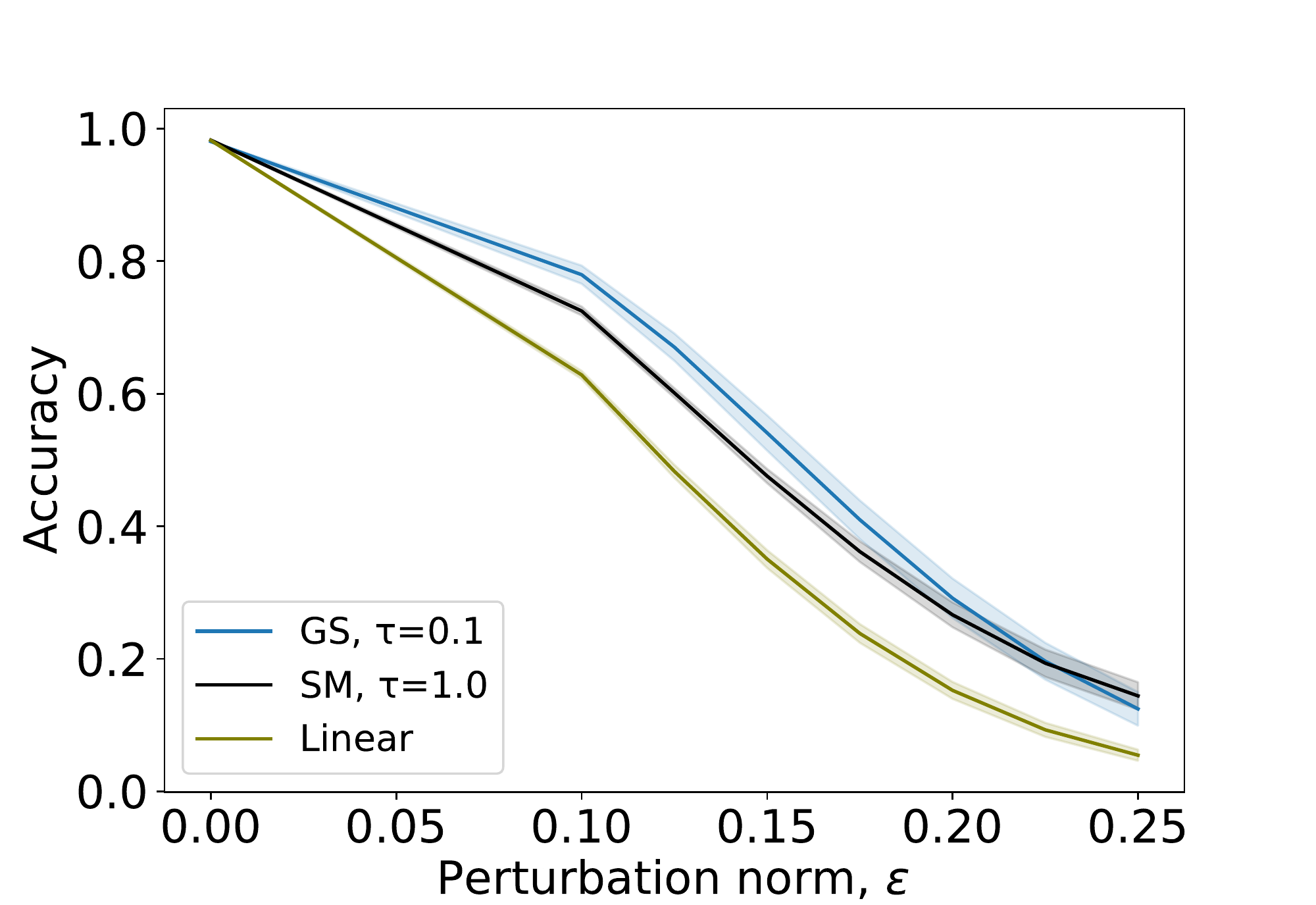}
\caption{Comparison to the baselines.}
  \label{fig:adversarial:b}
 \end{subfigure}
\caption{Robustness to adversarial examples: higher accuracy given fixed $\epsilon$ implies more robustness.}
\label{fig:adversarial}
\end{figure*}

Further, we hypothesize that the emergent discrete bottleneck might have useful properties, since 
 existing (continuous) architectures that explicitly impose a bottleneck pressure are more robust to
overfitting~\citep{Fischer2019} and adversarial
attacks~\citep{Alemi2016,Fischer2019}. We test  whether similar regularization properties also emerge in our computational
simulations (without any explicit pressure imposed through the cost function), and whether they are correlated with communication channel discreteness. If this connection exists, it  
also suggests that discreteness might be ``beneficial'' to
human languages for the same reasons.

\subsubsection{Robustness to over-fitting}

To assess our hypotheses, we consider the Image Classification game ($N_{l}=10$) in presence of randomly-shuffled training labels (the test set is untouched)~\citep{Zhang2016}.  This task allows us to explore whether the discrete communication bottleneck is associated to robustness to overfitting, and whether the latter depends on discreteness level (controlled by the temperature $\tau$ of Gumbel-Softmax). We use the same architecture as above. The agents are trained with Gumbel-Softmax relaxation; at test-time the communication is fully discrete. %

We also consider two baseline architectures without the discrete channel. In {Linear}, the fully connected output layer of Sender is directly connected to the linear embedding input of Receiver.
{Softmax} (SM) places a softmax activation (with temperature) after Sender's output layer and passes the result to Receiver. %

We vary temperature and proportion of training examples with shuffled  labels. We use temperatures $\tau = 1.0$ and $\tau = 10.0$ (the agents reach a test accuracy of 0.98 when trained with these temperatures on the original training set). SM with $\tau=1.0$ and $\tau=10.0$ behave similarly, hence we only report SM with $\tau=1.0$.

Figure~\ref{fig:broken_mnist:c} shows \emph{training} accuracy when all labels are shuffled.  Linear and SM fit the random labels almost perfectly within the first 150 epochs. With $\tau=10.0$, GS achieves 0.8 accuracy within 200 epochs. When GS with $\tau=1.0$ is considered, the agents only start to improve over random guessing after 150 epochs, and accuracy is well below 0.2 after 200 epochs. As expected, test set performance is at chance level (Figure~\ref{fig:broken_mnist:d}). 
In the next experiment, we shuffle labels for a randomly selected half of the training instances. Train and test accuracies are shown in Figures~\ref{fig:broken_mnist:e} and~\ref{fig:broken_mnist:f}, respectively. All models initially fit the true-label examples (train accuracy $\approx 0.5$, test accuracy $\approx 0.97$). With more training, the baselines and GS with $\tau=10.0$ start (over)fitting the random labels, too: train accuracy grows, while test accuracy falls.
In contrast, GS with $\tau=1.0$ does not fit random labels, and its test accuracy stays high. Note that SM patterns with Linear and high-temperature GS, showing that the training-time discretization
noise in GS is instrumental for robustness to over-fitting.

We interpret the results as follows. To fully exploit their joint capacity
for ``successful'' over-fitting, the agents need to coordinate label
memorization. This requires passing large amounts of information
through the channel. With a low temperature (more closely
approximating a discrete channel), this is hard, due to a stronger
entropy minimization pressure. To test the hypothesis, we run an
experiment where all labels are shuffled and a layer of size 400x400
is either added to Sender (just before the channel) or to Receiver
(just after the channel). We predict that, with higher $\tau$ (less
discrete, less entropy minimization pressure), the training curves
will be close, as the extra capacity can be used for memorization
equally easy in both cases.  With lower $\tau$ (more discrete, more pressure), the accuracy curves will be more
distant, as the extra capacity can only be successfully exploited for
memorization when placed \emph{before} the
channel. Figures~\ref{fig:broken_mnist:g} \& \ref{fig:broken_mnist:h}
bear out the prediction.

\subsubsection{Robustness to adversarial examples}
We study next robustness of agents equipped with a relaxed discrete channel against adversarial attacks. We use the same architectures as in the preceding experiment. %

We train agents with different random seeds and implement white-box attacks on the trained models, varying temperature $\tau$ and the 
allowed perturbation norm, $\epsilon$. We use the standard \textit{Fast Gradient Sign Method} of \cite{Goodfellow2014}. 
The original image $\vi_s$ is perturbed to $\vi^*_s$ along the direction that maximizes the loss of Receiver's output $\vo = R(S(\vi_s))$ 
w.r.t.\ the ground-truth class $\vl$:
\begin{equation}
    \vi_s^* = clip \left[ \vi_s + \epsilon \cdot sign \left[ \nabla_{\vi_s} \mathcal{L} (\vo, \vl) \right], 0, 1 \right],
\end{equation}
where $\epsilon$ controls the $L_{\infty}$ norm of the perturbation. 
Under an attack with a fixed $\epsilon$, a more robust method will have a higher accuracy. To avoid numerical stability issues akin to those reported by~\cite{Carlini2016}, all computations are done in 64-bit floats. 

We experiment with two approaches of getting gradients for the attack. Under the first approach, the gradient $\nabla_{\vi_s} \mathcal{L} (\vo, \vl)$ is estimated using the standard Gumbel-Softmax relaxation. It is possible, however, that the randomization that Gumbel-Softmax uses internally reduces the usefulness of gradients used for the attack. Hence we also experiment with a setup that is easier for an adversary: after training (and during the attack), we replace the Gumbel-Softmax by a softmax non-linearity with the same temperature. We found that performance in these two setups is virtually the same, indicating that the obtained robustness results are  independent from the randomization in the channel. Rather, they are due to emergence of well-separated ``categories'' during training.

As in the preceding experiment, SM behaves similarly with different temperatures (we experimented with $\tau \in \{0.1, 1.0, 10.0\}$): we only report results with $\tau=1.0$. Figure~\ref{fig:adversarial:a} shows that, as temperature decreases, the accuracy drop also decreases. The highest robustness is achieved with  $\tau=0.1$. 
Comparison with the baselines (Figure~\ref{fig:adversarial:b}) confirms that relaxed discrete training with $\tau=0.1$ improves~robustness.  %

{In sum, increased channel discreteness makes it harder to transmit large amounts of information, and leads to increased robustness against over-fitting and adversarial examples. Discreteness brings about a bottleneck that has beneficial properties, which might ultimately provide a motivation for why an emergent communication system should  evolve towards discreteness.
}
\section{Related Work}
\label{sRelated}

We briefly reviewed studies of emergent deep agent
communication and  entropy minimization in human language in the
introduction. We are not aware of earlier work that looks for this
property in emergent communication, although \citet{Evtimova:etal:2018}
used information theory to study protocol development during learning,
and, closer to us, \citet{Kaageback:etal:2018} studied the effect of
explicitly adding a complexity minimization term to the cost function
of an emergent color-naming
system.

Discrete representations are explored in many
places~\citep[e.g.,][]{Oord2017,Jang2016,Rolfe2016}.  However, these works focus on ways to learn good discrete
representations, rather than analyzing the
properties of representations that are independently emerging on the
side. { Furthermore, our study extends to agents communicating with variable-length messages, produced and consumed by GRU~\citep{Cho2014} and Transformer~\citep{Vaswani:etal:2017} cells (see Supplementary). The sequential setup is specific to language, clearly distinguished from the settings studied in generic sparse-representation work.}

Other studies, inspired by the Information Bottleneck principle, control the complexity of neural  representations by regulating their information content~\citep{Strouse2017,Fischer2019,Alemi2016,Achille2018,Achille2018b}. While they externally impose the bottleneck, we observe that the latter is an intrinsic feature of learning to communicate through a discrete channel.

\section{Discussion}
\label{sDiscussion}

Entropy minimization is pervasive in human language, where it
constitutes a specific facet of the more general pressure towards
communication efficiency. We found that the same property consistently
characterizes the protocol emerging in simulations where two neural
networks learn to solve a task jointly through a discrete
communication code.

In a comparative perspective, we hypothesize that entropy
minimization is a general property of discrete
communication, independent of specific biological
constraints humans are subject to. In particular, our analysis
tentatively establishes a link between this property and the inherent
difficulty of encoding information in discrete form (cf.~the effect of
adding a layer before or after the communication bottleneck in the
over-fitting experiment).

Exploring entropy minimization in computational simulations
provides a flexibility we lack when studying humans. For example, we
uncovered here initial evidence that the communication bottleneck is
acting as a good regularizer, making the joint agent system more
robust to noise and adversarial examples. This leads to an intriguing conjecture on the origin
of language. Its discrete nature is often
traced back to the fact that it allows us to produce an infinite
number of expressions by combining a finite set of primitives
\citep[e.g.,][]{Berwick:Chomsky:2016}. However, it is far from clear
that the need to communicate an infinite number of concepts could have
provided the initial pressure to develop a discrete code. More
probably, \emph{once such code independently emerged}, it laid the conditions to develop an infinitely expressive language
\citep{Bickerton:2014,Collier:etal:2014}. Our work suggests that,
because of its inherent regularizing effect, discrete coding is advantageous
already when communication is about a limited number of concepts, providing an alternative explanation for its origin.

In the future, we would like to study
more continuous semantic domains, such as color maps, where perfect accuracy is
not easily attainable, nor desirable. Will the networks find an
accuracy/complexity trade-off similar to those attested in human
languages? Will other core language properties claimed to be
related to this trade-off, such as Zipfian frequency
distributions \citep{FerrerICancho:DiazGuilera:2007}, concurrently emerge? We would also like to compare the performance of human subjects equipped with novel continuous vs.~discrete communication protocols, adopting the methods of experimental semiotics \citep{Galantucci:2009}. We expect discrete protocols to be more general and robust.

Our results have implications for the efforts to evolve agents
interacting with each other and with humans through a discrete
channel. First, because of entropy minimization, we should not
 agents to develop a richer protocol than the simplest one
 ensuring accurate communication. For example,
\citet{Bouchacourt2018} found that agents trained to discriminate pairs
of natural images depicting instances of about 500 high-level
categories, such as cats and dogs, developed a lexicon that does not
denote such categories, but low-level properties of the images
themselves. This makes sense from an entropy-minimization
perspective, as talking about the 500 high-level categories demands
$\log_2 500$ bits of information, whereas many low-level strategies
(e.g., discriminating average pixel intensity in the images) will only
require transmitting a few bits. To have agents developing rich
linguistic protocols, we must face them with varied challenges
that truly demand them.

Second, the focus on a discrete protocol is typically motivated by the goal to
develop machines eventually able to communicate with
humans. Indeed, discrete messages are not required in multi-agent
scenarios where no human in the loop is foreseen
\citep{Sukhbaatar2016}. Our results suggest that, long
before agents reach the level of complexity necessary to converse with
humans, there are independent reasons to encourage discreteness, as it
leads to simpler protocols and it provides a source of robustness in a noisy world. An exciting
direction for future applied work will be to test %
the effectiveness of discrete communication as a general
form of representation learning.

\textbf{Acknowledgements} The authors thank Emmanuel Dupoux for discussions and the anonymous reviewers for their feedback.

\bibliography{other,biblio}
\bibliographystyle{icml2020}

\section{How much does Receiver rely on messages in Guess Number?}
We supplement the experiments of Section~3 of the main text by studying the degree to which Receiver relies on messages in Guess Number. In particular, we show that when Receiver has the full input ($\vi_s = \vi_r$), it ignores the messages.

We measure the degree to which Receiver relies on the messages from Sender by constructing a setup where we break communication, but still let Receiver rely on its own input. More precisely, we first enumerate all test inputs for Sender $\vi_s$ and Receiver $\vi_r$. We obtain messages that correspond to Sender's inputs, and shuffle them. Next, we feed the shuffled messages alongside Receiver's own (un-shuffled) inputs and compute accuracy, as a measure of Receiver's dependence on the messages. This procedure preserves the marginal distribution of  Sender's messages, but destroys all the information Sender transmits.

\begin{figure}
  \center
    \begin{subfigure}{1.0\linewidth}
    \includegraphics[width=0.9\linewidth]{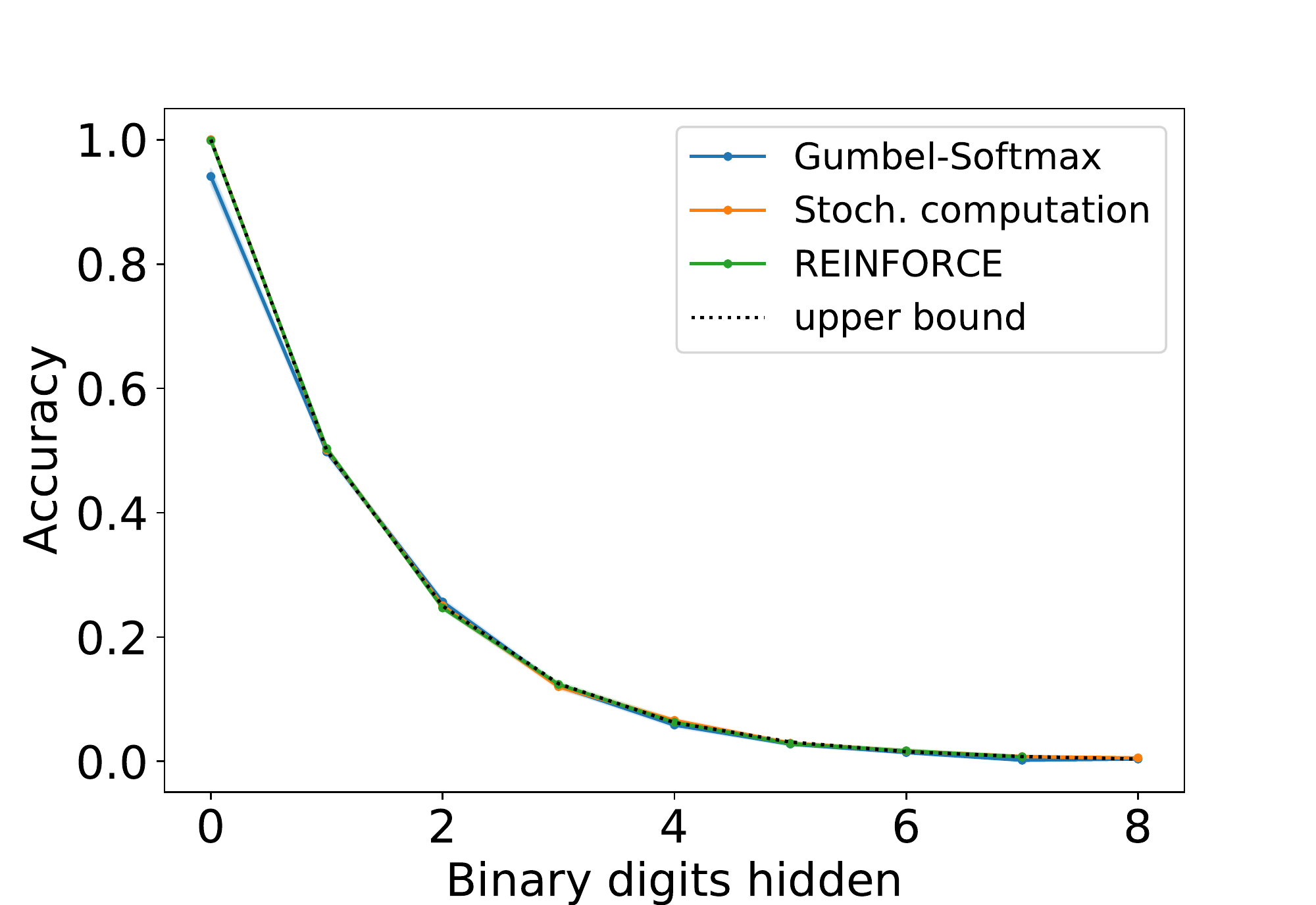}
    \end{subfigure}%
    \caption{Guess Number: Receiver's dependence on messages, measured as performance drop under message intervention.}
    \label{shuffle:guess}
\end{figure}

Without messages, Receiver, given  $k$ input bits, can only reach an accuracy of $2^{8 - k}$. In Figure~\ref{shuffle:guess}, we report results aggregated by training method. Receiver is extremely close to the accuracy's higher bound in all configurations. Moreover, when Receiver gets the entire input, the drop in accuracy  after  shuffling is tiny, proving that Receiver's reliance on the message is minimal in that setting.

\section{Influence of architecture choices}
\subsection{Does vocabulary size affect the results?}
\begin{figure*}
  \centering
\begin{subfigure}{0.33\linewidth}
  \centering
    \includegraphics[width=1.0\linewidth]{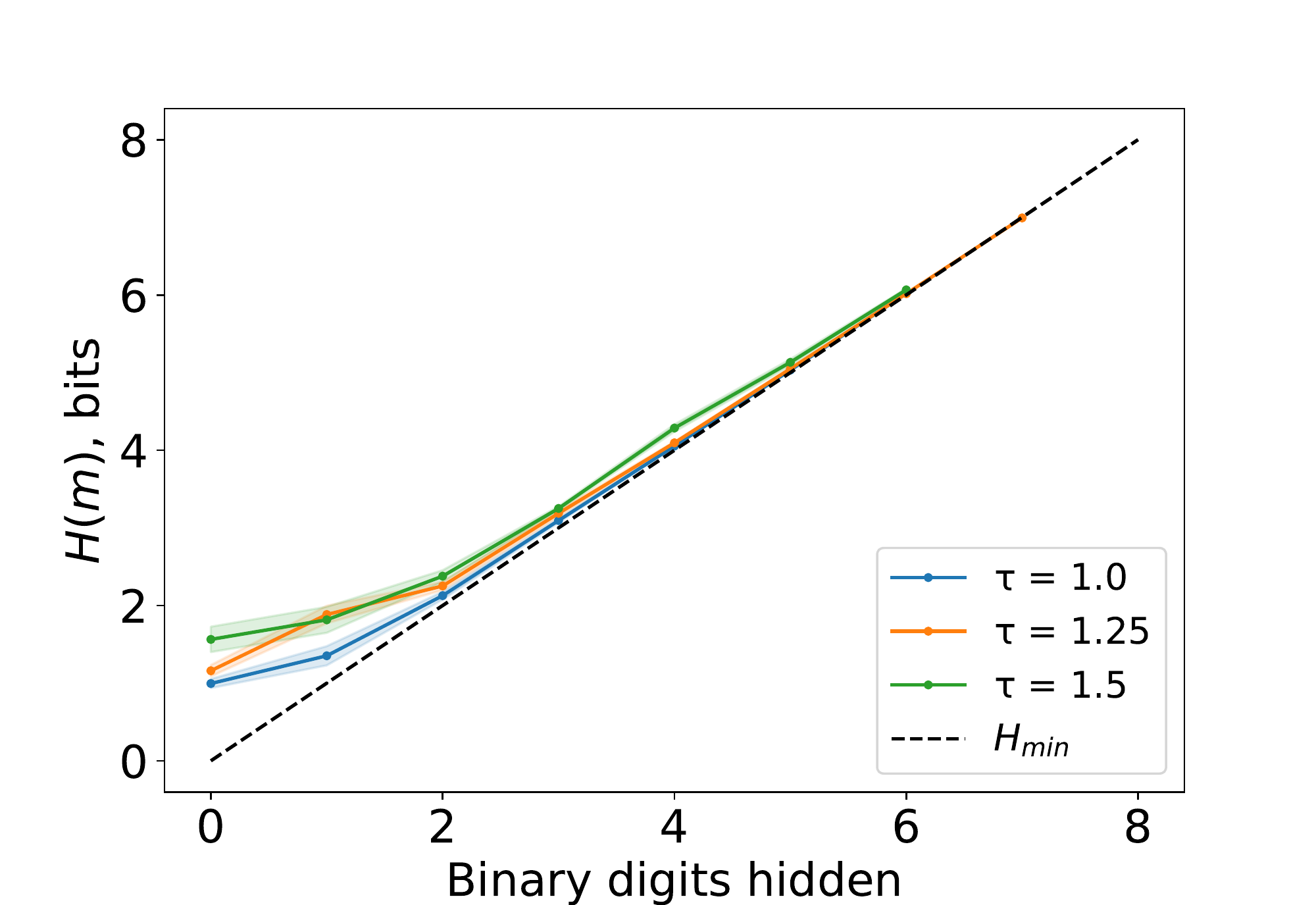}
  \caption{Vocab. size: 256, Gumbel-Softmax}
\end{subfigure}%
\begin{subfigure}{0.33\linewidth}
  \centering
    \includegraphics[width=1.0\linewidth]{pics/guess/hot_cold_info_1024.pdf}
  \caption{Vocab. size: 1024, Gumbel-Softmax}
\end{subfigure}%
\begin{subfigure}{0.33\linewidth}
  \centering
    \includegraphics[width=1.0\linewidth]{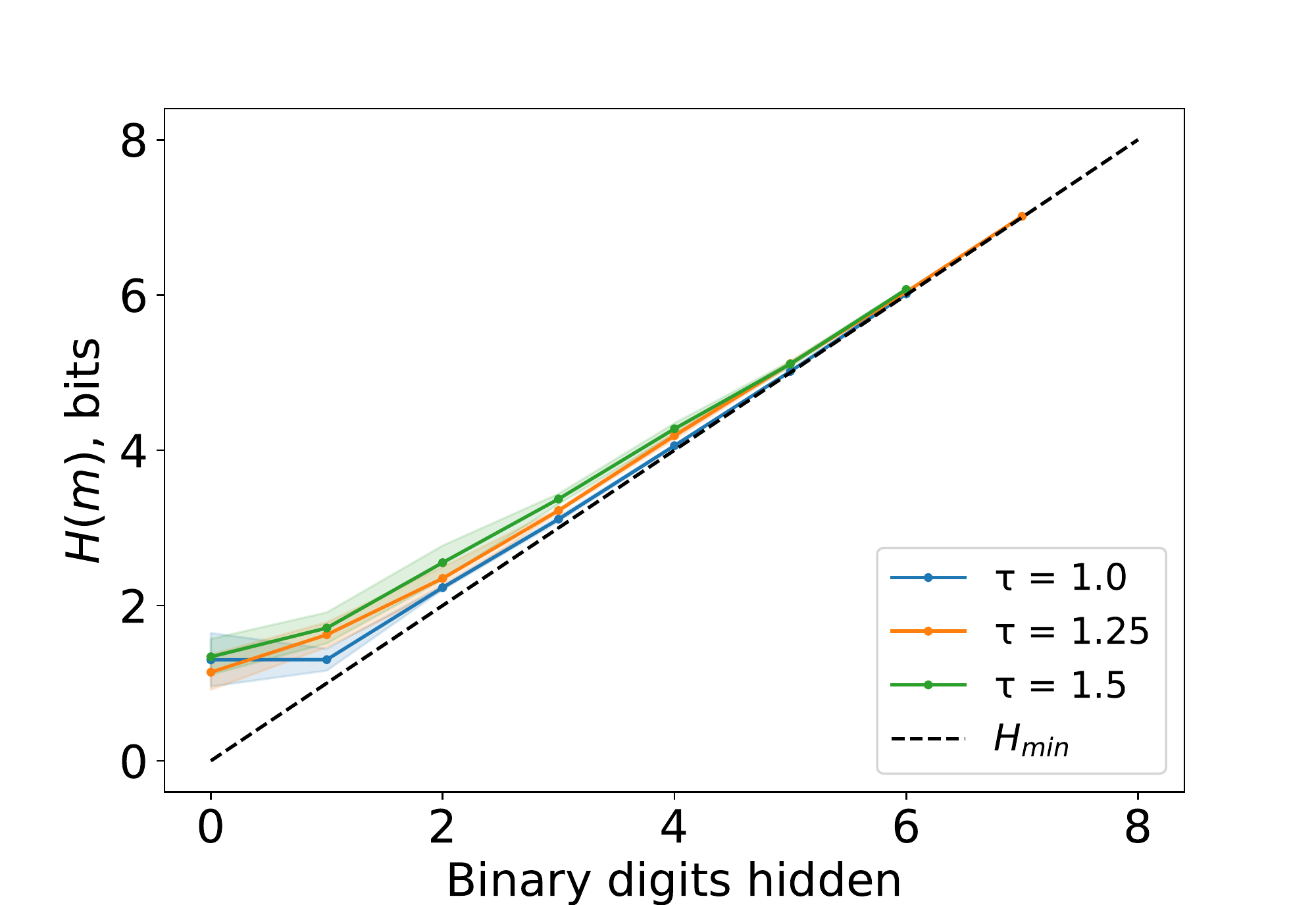}
  \caption{Vocab. size: 4096, Gumbel-Softmax}
\end{subfigure}

\begin{subfigure}{0.33\textwidth}
  \centering
  \includegraphics[width=1.0\textwidth]{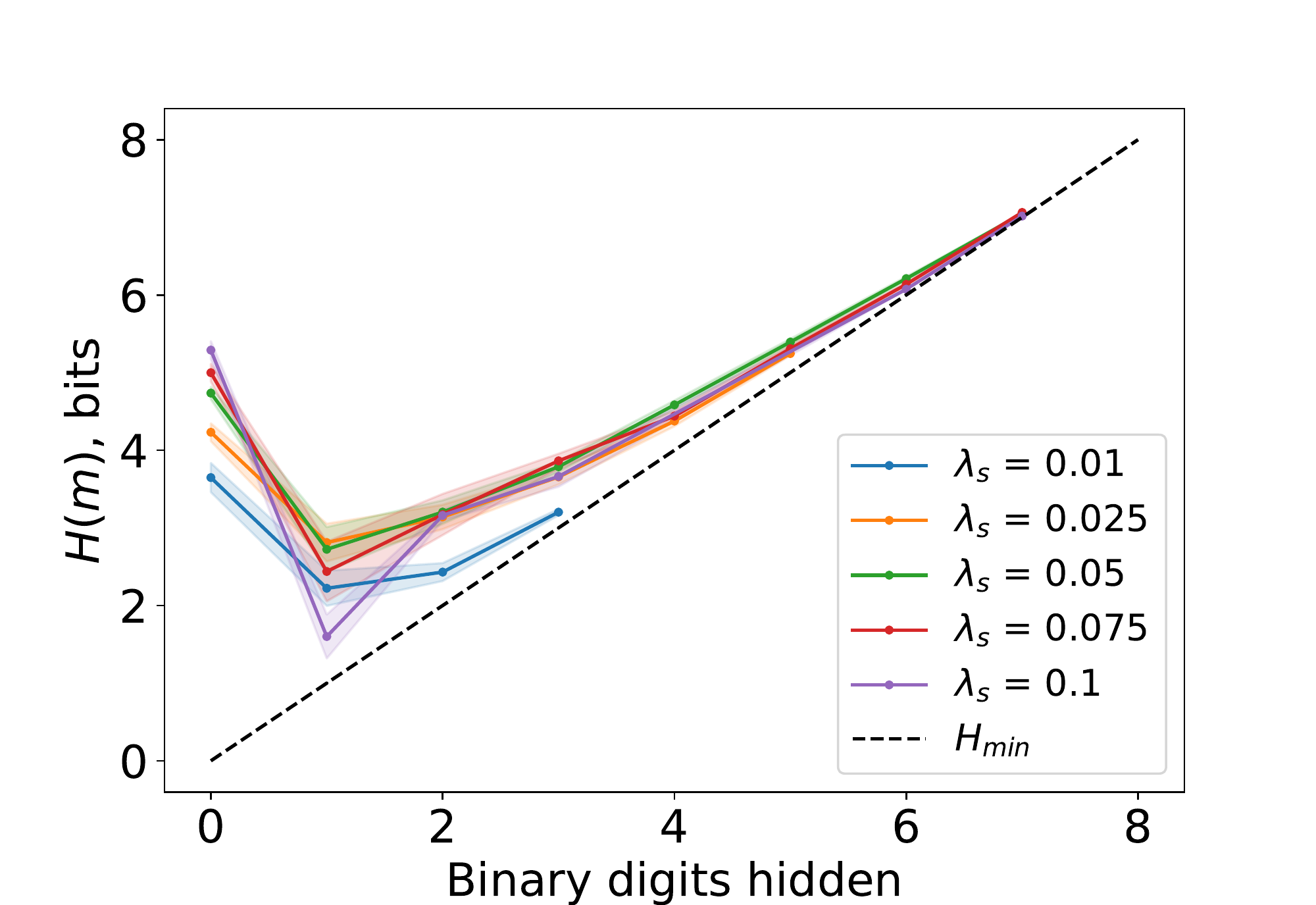}
  \caption{Vocab. size: 256, Stoch.~Computation Graph approach}
\end{subfigure}%
\begin{subfigure}{0.33\textwidth}
  \centering
  \includegraphics[width=1.0\textwidth]{pics/guess/rf_by_l_1024.pdf}
  \caption{Vocab. size: 1024, Stoch.~Computation Graph approach}
\end{subfigure}%
\begin{subfigure}{0.33\textwidth}
  \centering
  \includegraphics[width=1.0\textwidth]{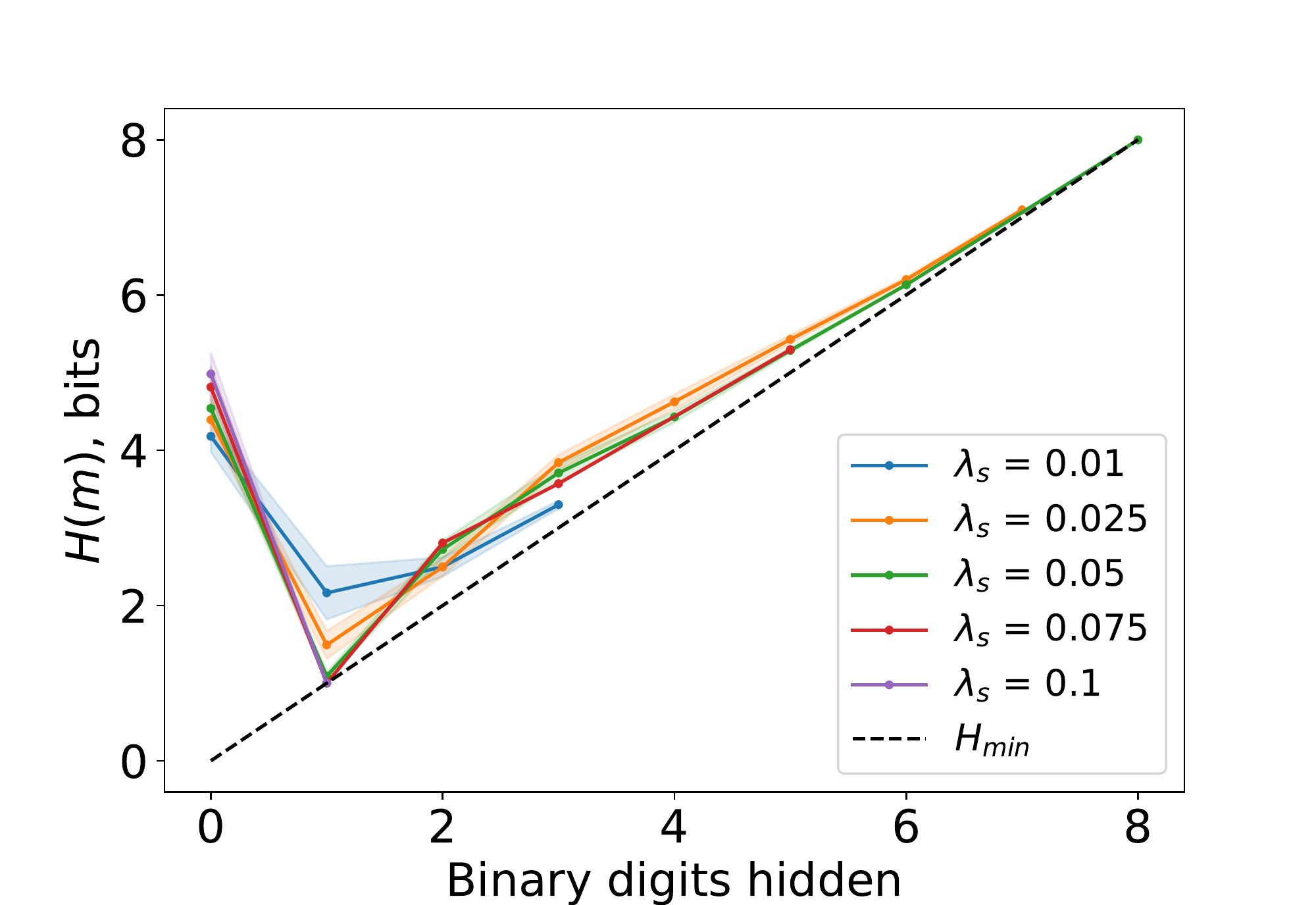}
  \caption{Vocab. size: 4096, Stoch.~Computation Graph approach}
\end{subfigure}

\begin{subfigure}{0.33\textwidth}
  \centering
  \includegraphics[width=1.0\textwidth]{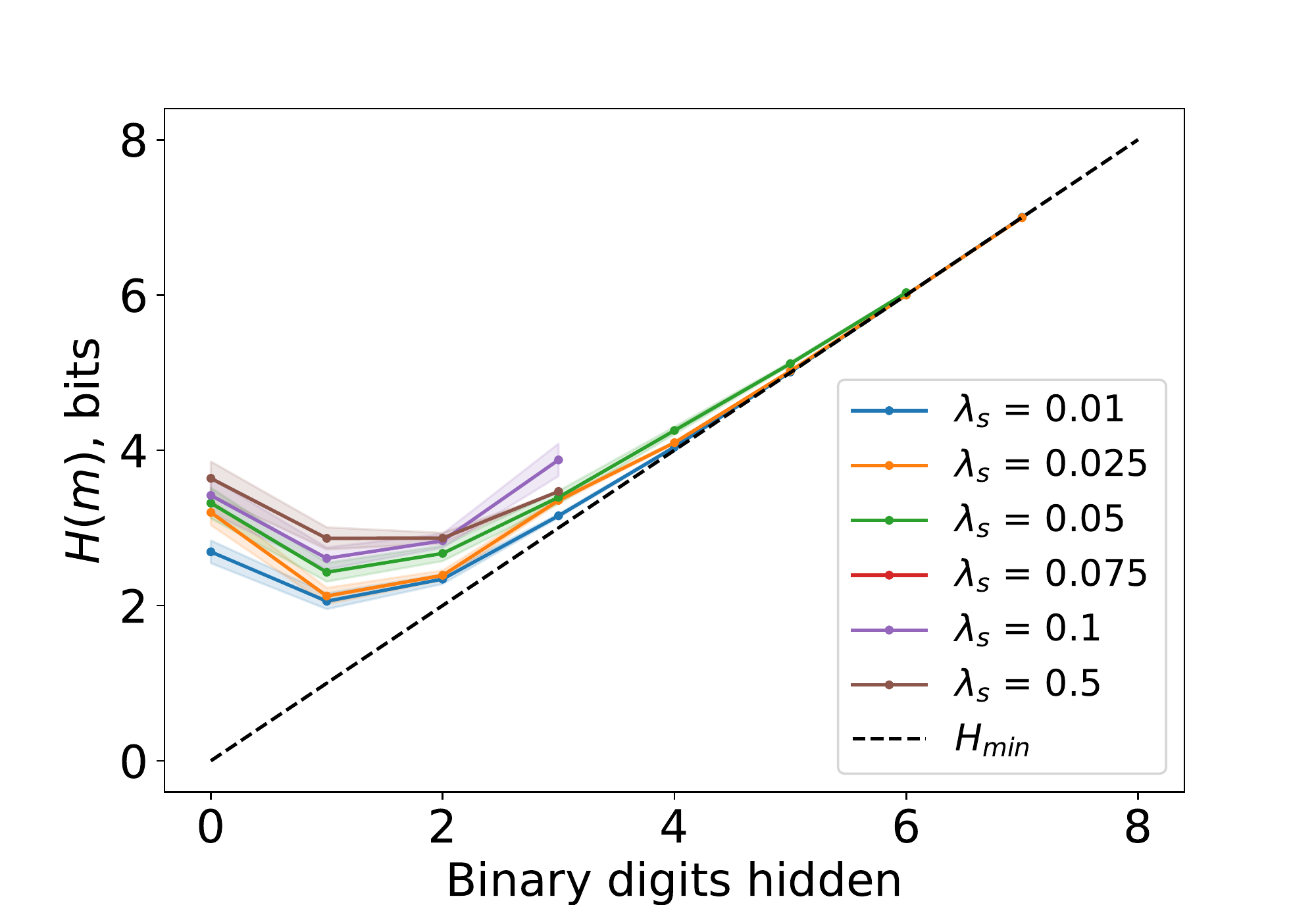}
  \caption{Vocab. size: 256, REINFORCE}
\end{subfigure}%
\begin{subfigure}{0.33\textwidth}
  \centering
  \includegraphics[width=1.0\textwidth]{pics/guess/pure_by_l_1024.pdf}
  \caption{Vocab. size: 1024, REINFORCE}
\end{subfigure}%
\begin{subfigure}{0.33\textwidth}
  \centering
  \includegraphics[width=1.0\textwidth]{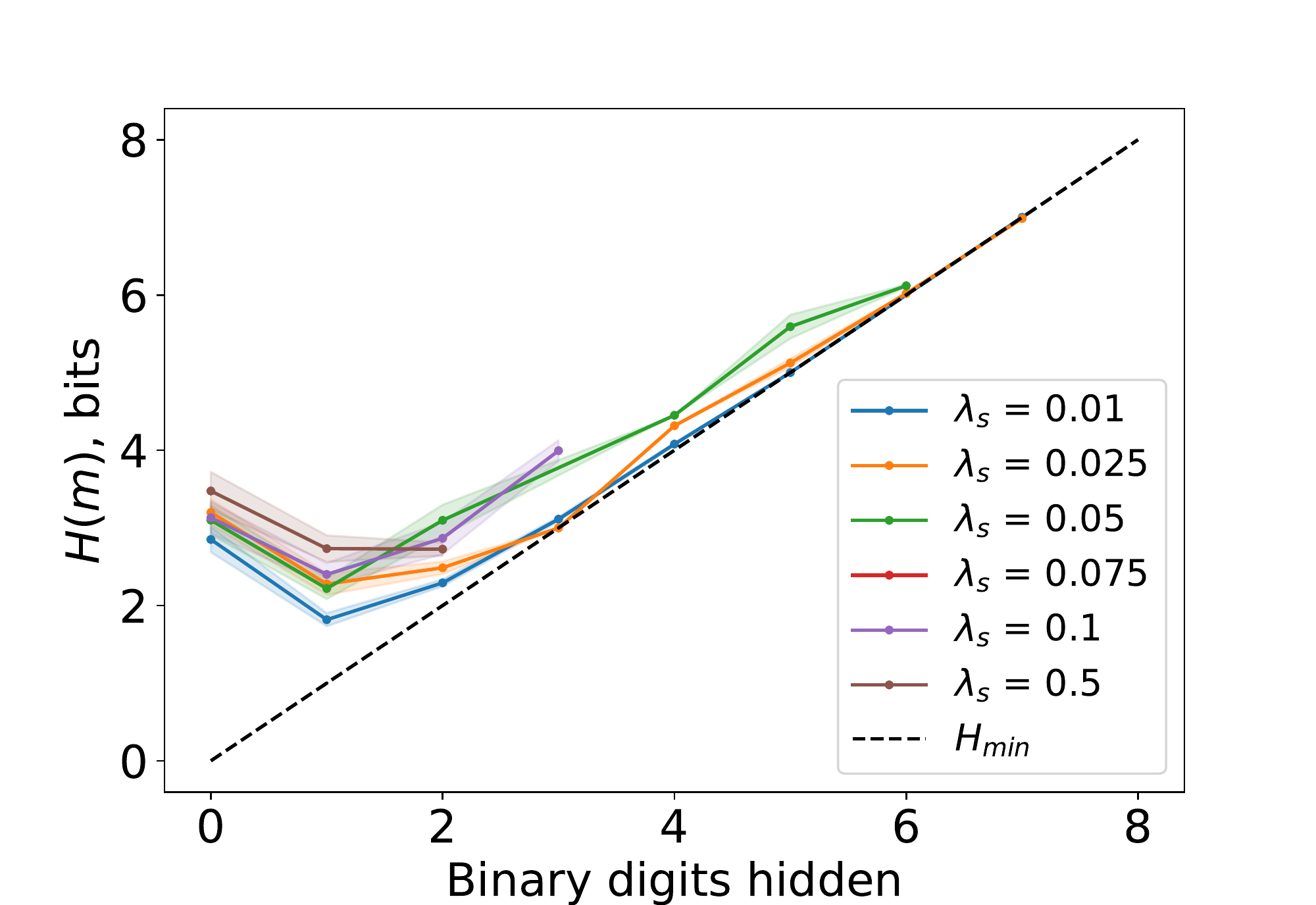}
  \caption{Vocab. size: 4096, REINFORCE}
\end{subfigure}
    \caption{Guess Number: Entropy of the messages $\vm$, depending on vocabulary size, training method, and relaxation temperature $\tau$ (when trained with Gumbel-Softmax) or Sender's entropy regularization coefficient $\lambda_s$. Shaded regions mark standard deviation.}
\label{fig:guess}
\end{figure*}

\begin{figure*}
  \centering
\begin{subfigure}{0.33\linewidth}
  \centering
  \includegraphics[width=1.0\textwidth]{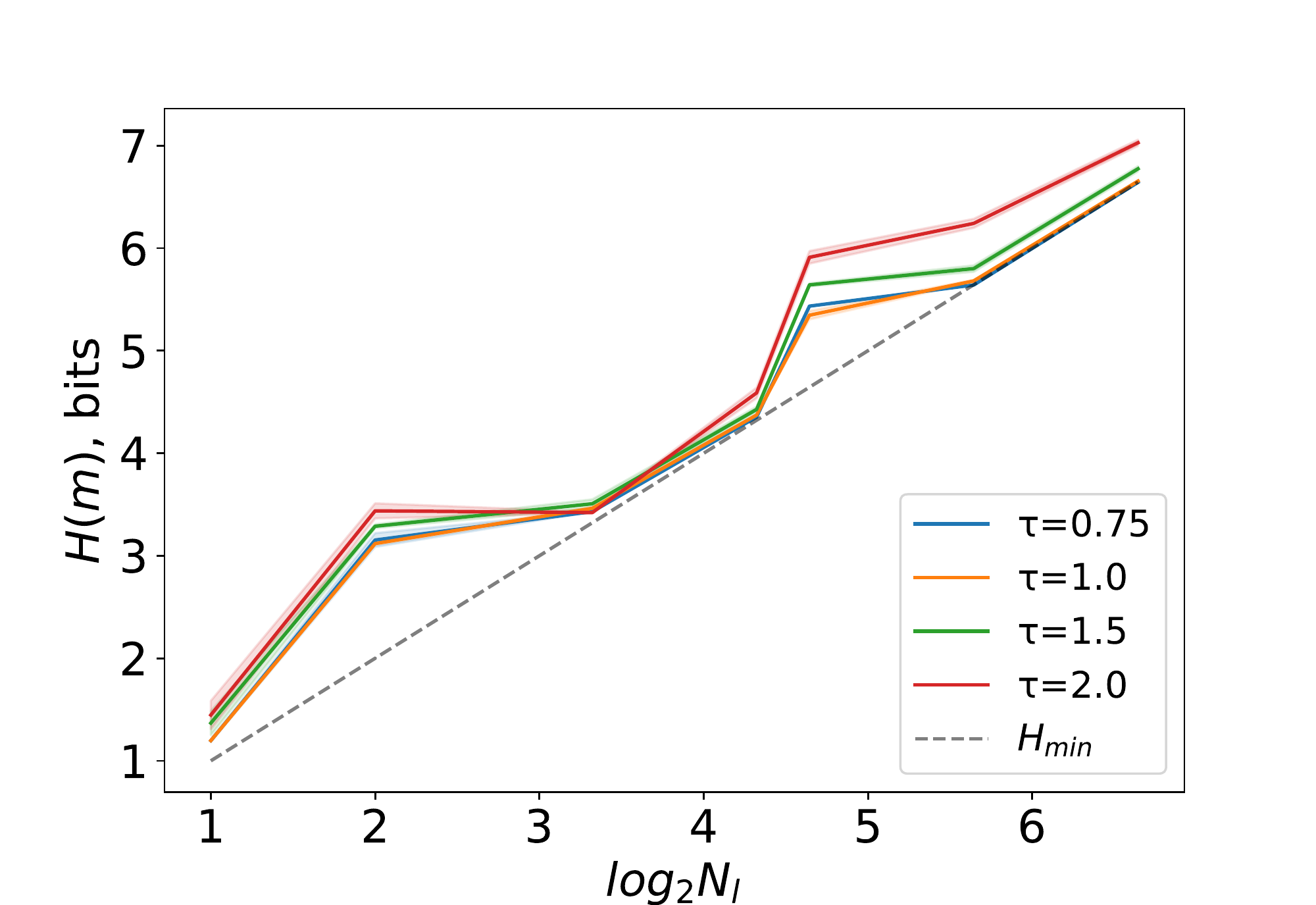}
  \caption{Vocab. size: 512}
\end{subfigure}%
\begin{subfigure}{0.33\linewidth}
  \centering
  \includegraphics[width=1.0\textwidth]{pics/mnist_classification/info_by_t_1024_0.pdf}
  \caption{Vocab. size: 1024}
\end{subfigure}%
\begin{subfigure}{0.33\linewidth}
  \centering
  \includegraphics[width=1.0\textwidth]{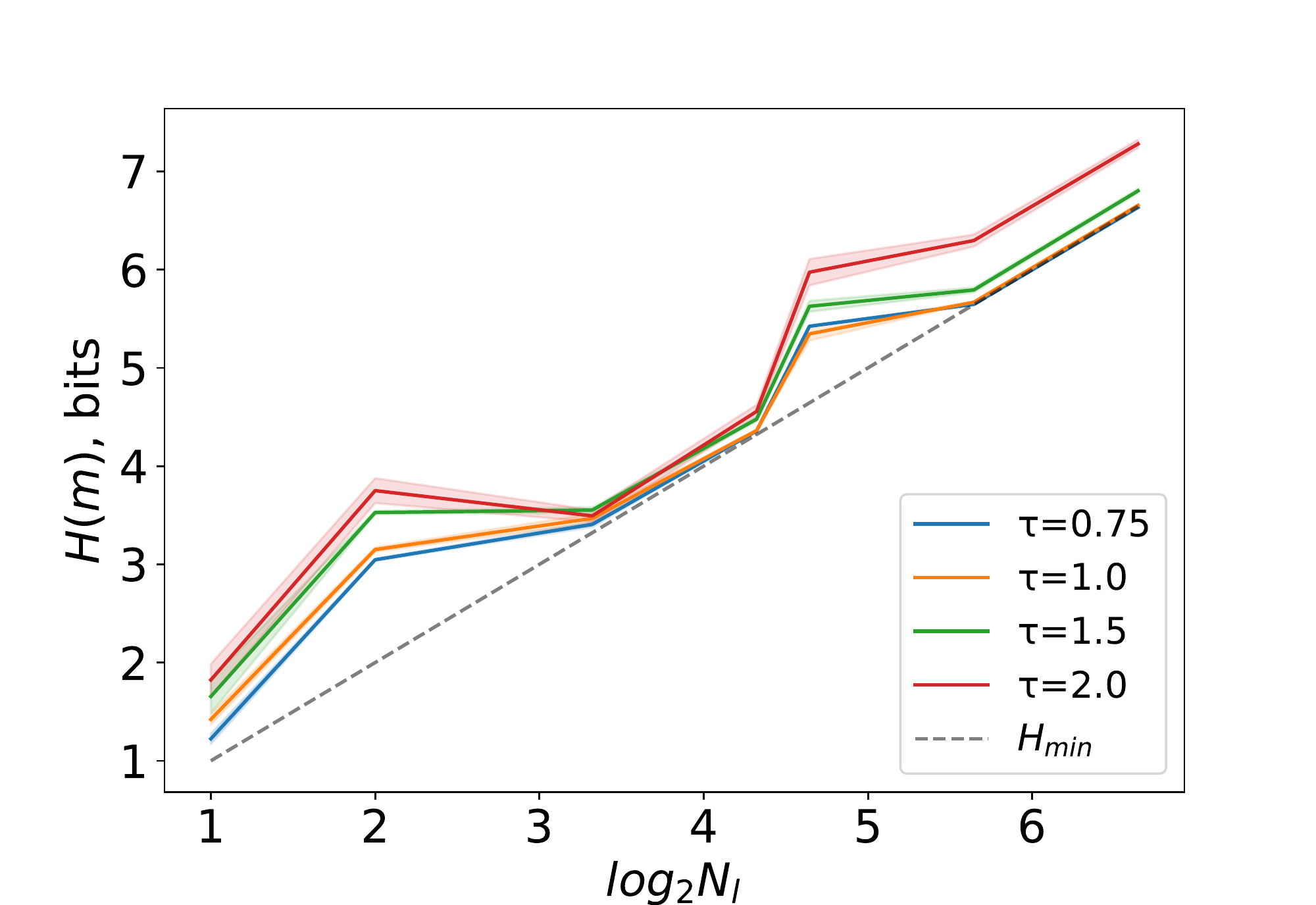}
\caption{Vocab. size: 2048}
\end{subfigure}%
\caption{Image Classification: entropy of the messages $H(\vm)$ across vocabulary sizes. Successful runs are pooled together.  Shaded regions mark standard deviation.}
\label{fig:mnist_vocab_t}
\end{figure*}

We repeat the same experiments as in Section~3 of the main text while varying vocabulary size. Note that, to make Guess Number solvable across each configuration, the vocabulary has to contain at least 256 symbols. Similarly, for Image Classification, vocabulary size must be of at least 100. We tried vocabulary sizes of 256, 1024, 4096 for Guess Number, and 512, 1024, 2048 for Image Classification. 
The results are reported in Figures~\ref{fig:guess} (Guess Number) and~\ref{fig:mnist_vocab_t} (Image Classification). We observe that there is little qualitative variation over vocabulary size, hence the conclusions we had in Section~3 of the main paper are robust to variations of this parameter.

\subsection{Does Receiver's capacity affect the results?}
One potential confounding variable is the capacity of Receiver. Indeed, if Receiver is very simple, then, for the task to be solved, Sender would have to calculate the answer itself and feed it to Receiver. 
To investigate this, we repeat the Image Classification experiment from Section~4.1 of the main paper while controlling the power of Receiver's architecture:  we put two additional fully-connected 400x400 hidden layers between the input embedding and the output layer, 
while in Section~4, Receiver had a single hidden layer.

\begin{figure}
  \centering
\begin{subfigure}{0.5\textwidth}
  \centering
  \includegraphics[width=1.0\textwidth]{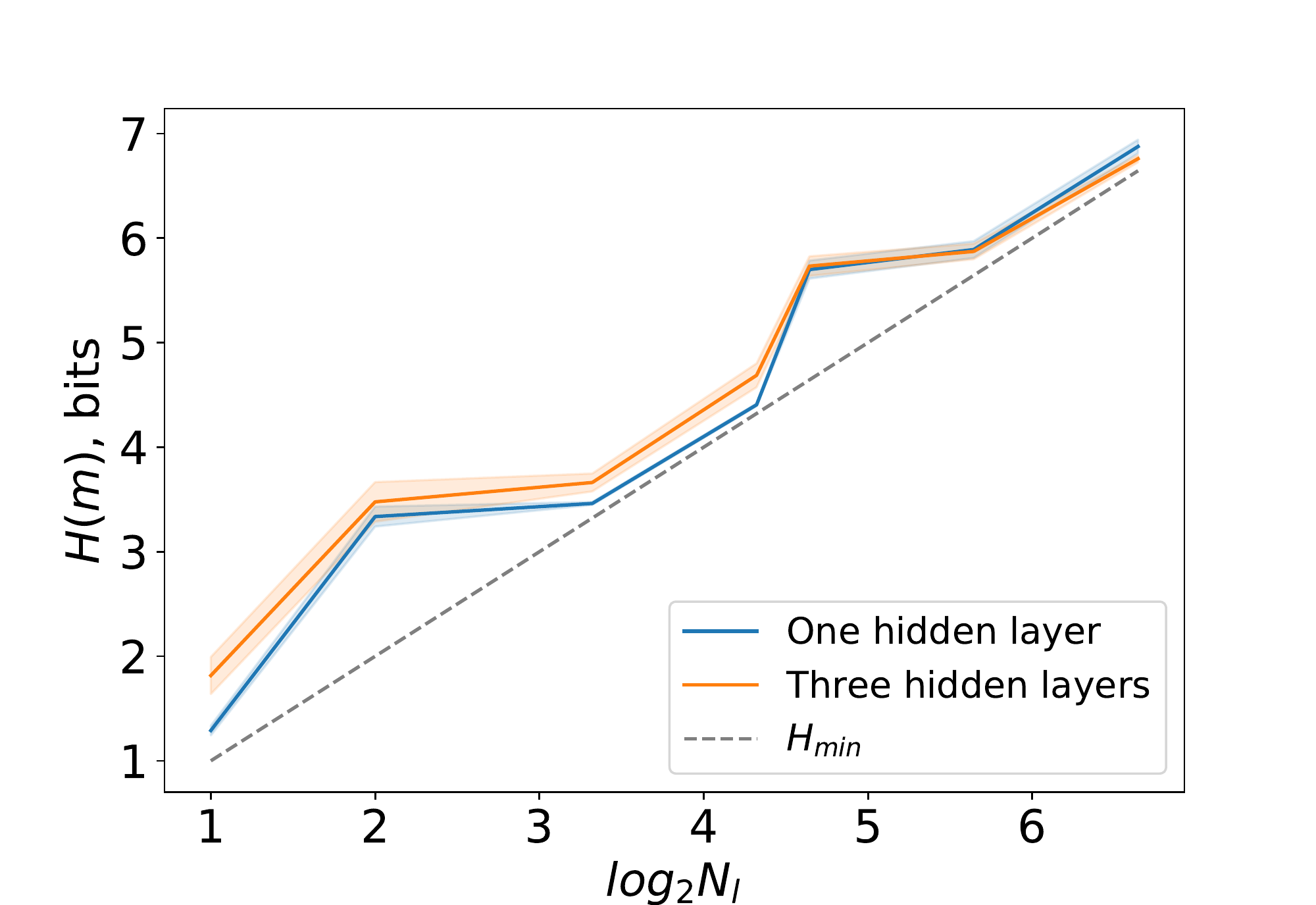}
\end{subfigure}
\caption{Image Classification: entropy of the messages $H(\vm)$ across Receiver model sizes. Successful runs are pooled together.  Shaded regions mark standard deviation.}
\label{fig:mnist_capacity}
\end{figure}

In Figure~\ref{fig:mnist_capacity}, we compare the results obtained with these two variations of Receiver. The reported entropy minimization effect holds: even in presence of additional layers, the entropy of messages $H(\vm)$ is far from the upper-bound $H_{max}=10$ bits and closely follows the lower bound, $H_{min} = \log_2 N_l$. Thus, again, a more nuanced protocol only appears when it is needed. Finally, we see that results for both architectures are close, although in three out of seven task setups (the number of classes $N_l$ is 2, 10, and 20) a deeper model results in a slightly higher entropy of the protocol, on average. Overall, we conclude that Receiver's capacity does not play a major role in the entropy minimization effect and the latter also takes place with a more powerful Receiver.

\subsection{What if communication takes place through sequences of symbols?}
We also experiment with Guess Number in a setup where the agents communicate via variable-length messages. The general architecture of the agents is same as in 
Section~3, but we append GRU agents~\citep{Cho2014}. 
Sender GRU is unrolled to generate the message. The message is produced until the GRU outputs a special \textit{eos} token or until the maximal length is reached. In the latter case, \textit{eos} is appended to the message. The produced message is consumed by a Receiver's GRU unit and the hidden state corresponding to \textit{eos} is used by Receiver as input to further processing. When Receiver has additional inputs (in the Guess Number game), these inputs are used as initial hidden state of the GRU cell. We use the Stochastic Computation Graph estimator as described in Section~3.2, as it provided fastest convergence.

\begin{figure*}
\centering
\begin{subfigure}{0.4\linewidth}
\centering
  \includegraphics[width=1.0\linewidth]{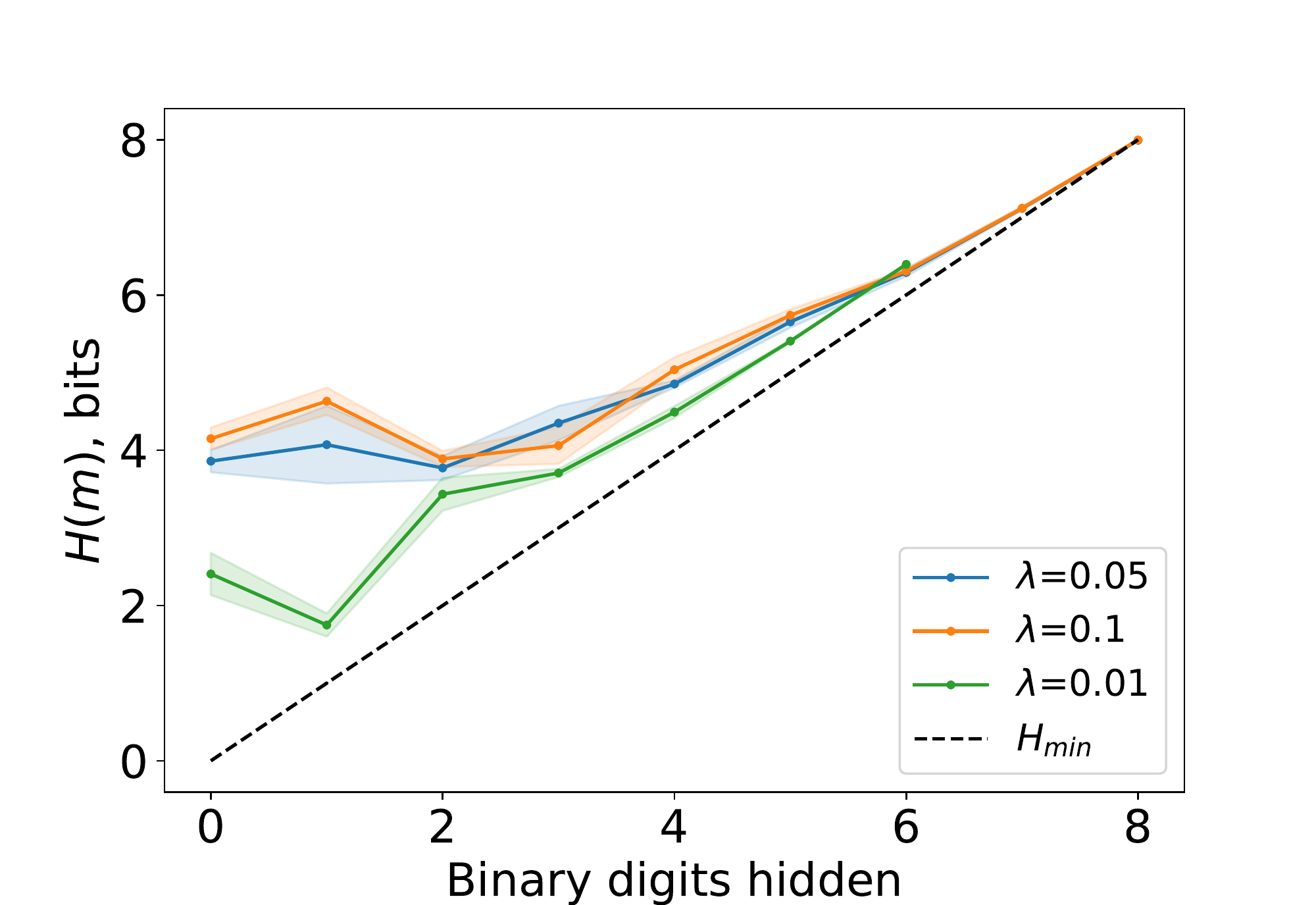}
\caption{Max length: 5, vocabulary size: 16}
 \end{subfigure}%
\begin{subfigure}{0.4\linewidth}
\centering
  \includegraphics[width=1.0\linewidth]{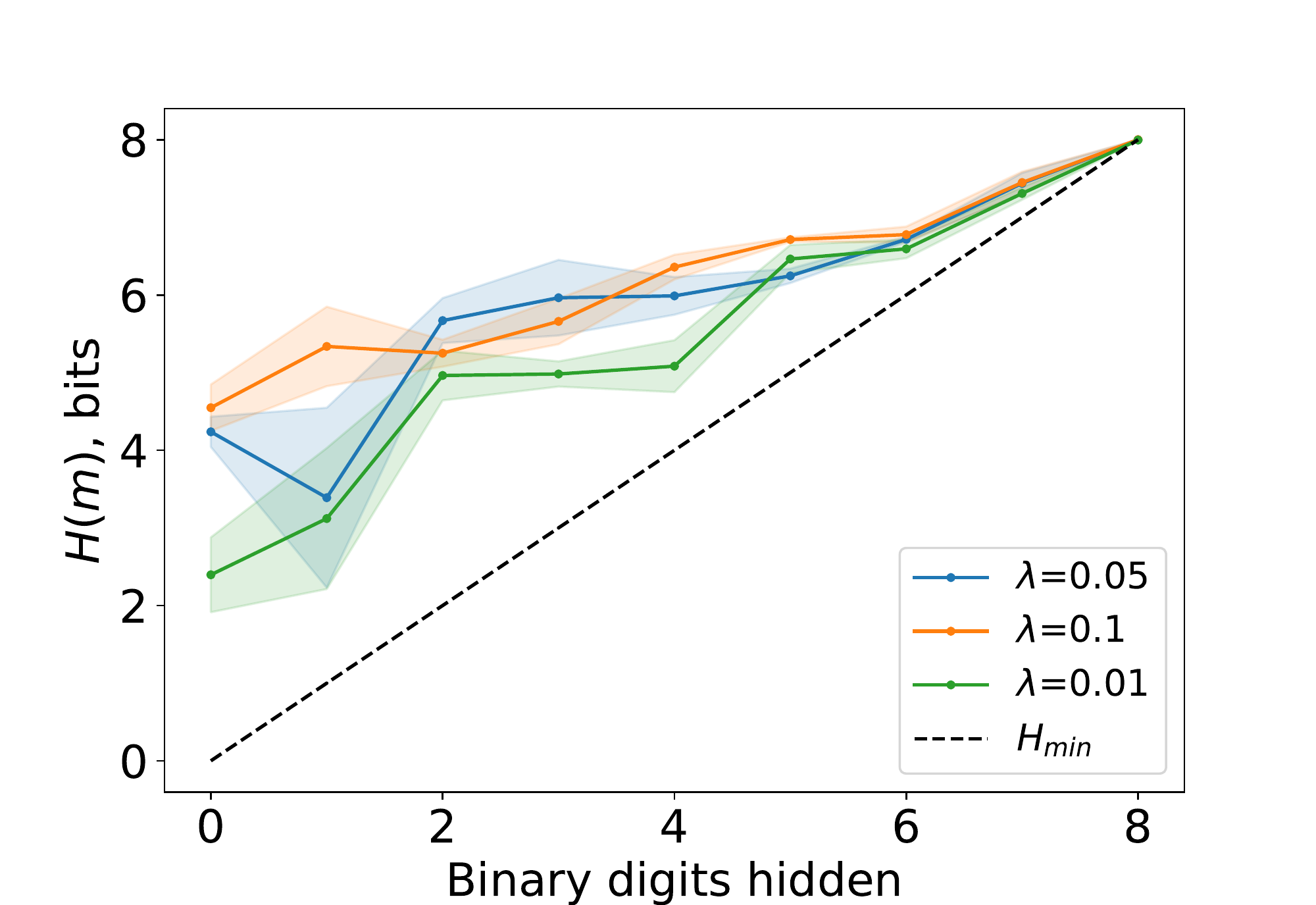}
\caption{Max length: 10, vocabulary size: 16}
 \end{subfigure}

 \begin{subfigure}{0.4\linewidth}
\centering
  \includegraphics[width=1.0\linewidth]{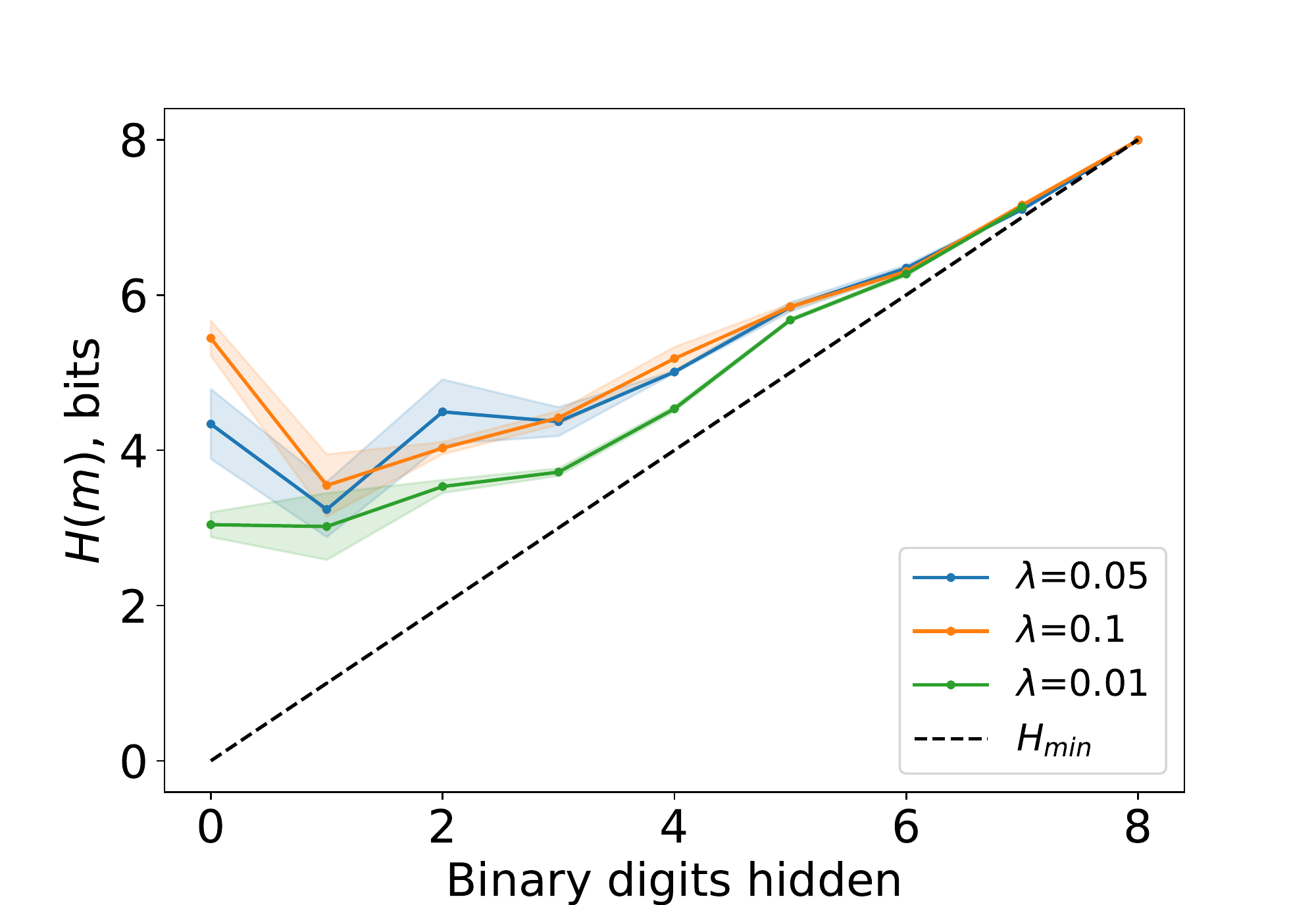}
\caption{Max length: 5, vocabulary size: 64}
 \end{subfigure}%
\begin{subfigure}{0.4\linewidth}
\centering
  \includegraphics[width=1.0\linewidth]{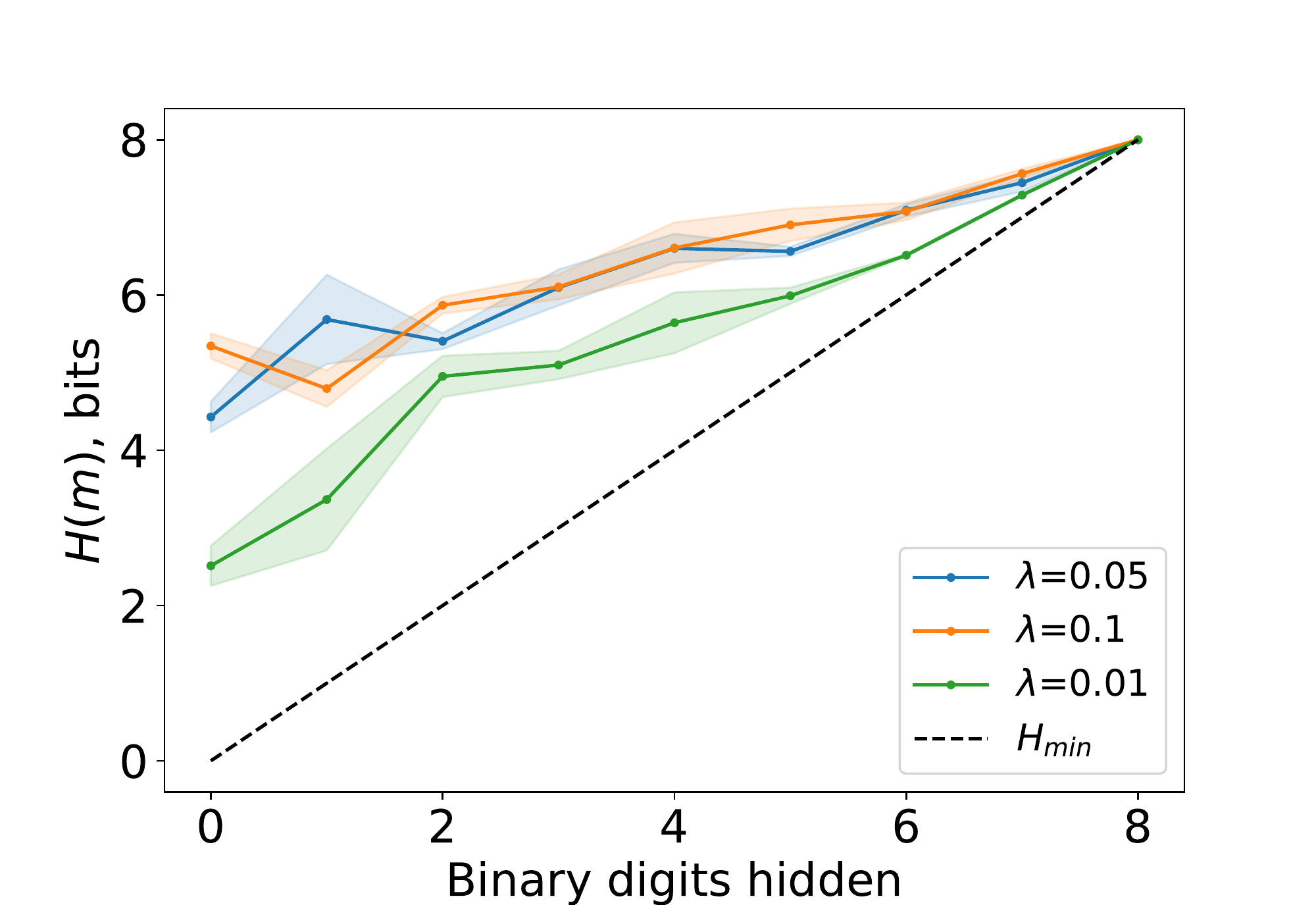}
\caption{Max length: 10, vocabulary size: 64}
\end{subfigure}
\caption{Guess Number: Entropy of the emergent protocol when communication is performed with variable-length messages. Shaded regions mark standard deviation.}
\label{fig:gs_varlen}
\end{figure*}

We consider the entire variable-length message as the realization of a random variable $\vm$ when calculating the entropy of the messages, $H(\vm)$. 
The results are reported in Figure~\ref{fig:gs_varlen}, arranged in function of maximal message length and vocabulary size. As before, we aggregate the successful runs according to the entropy regularization coefficient 
$\lambda_s$ applied to Sender's output layer.

From Figure~\ref{fig:gs_varlen} we observe that the results are in line with those obtained in the one-symbol scenario. Entropy minimization still holds: a more nuanced (high-entropy) protocol only develops when more digits are hidden from Receiver, which hence requires more information to perform the task. 
The approximation to the lower bound is however less tight as the overall number of possible messages grows (higher maximum length and/or vocabulary size). There is also a weak tendency for lower $\lambda_s$ to encourage a tighter bottleneck.

In preliminary experiments, we have similar results when the variable-length communication is performed via Transformer cells~\citep{Vaswani:etal:2017} instead of GRUs (not reported here).

\section{Two-digit MNIST dataset}
As discussed in Section~3, to ensure high output informational complexity in the Image Classification task, we use a two-digit variant of the MNIST dataset \citep{LeCun:etal:1998}. We construct it as follows. When iterating over the original MNIST dataset, we take a batch $b$ and (a) select the first $|b| / 2$ and last $|b| / 2$ images, refer to them as $b_1$ and $b_2$, respectively; (b) create a new batch where the $i$th image from $b_1$ is placed to the left of the $i$th image from $b_2$ and then \emph{vice versa}. As a result, we obtain a new stream of images, where each MNIST digit is seen twice, on the left and on the right side. Note that not all possible pairwise combinations of the original images are generated (there are $60000^2$ of those in the training set alone) and the exact combinations change across epochs. As labels, we use the depicted two-digit number modulo $N_l$, where $N_l$ is the required number of classes. All pixels are scaled into {[}0, 1{]}. We use this same process to generate training and test sets, based on the training and test images of the original MNIST dataset, respectively.

\section{Hyperparameters}
In our experiments, we used the following hyperparameter grids.

\textbf{Guess Number (Gumbel-Softmax)} 
Vocab.\ size: {[}256, 1024, 4096{]}; temperature, $\tau$: {[}0.5, 0.75, 1.0, 1.25, 1.5{]}; learning rate:  {[}0.001, 0.0001{]};  max.\ number of epochs: 250; random seeds: {[}0, 1, 2, 3{]}; batch size: 8; early stopping thr.: 0.99; bits shown to Receiver: {[}0, 1, 2, 3, 4, 5, 6, 7, 8{]}.

\textbf{Guess Number (REINFORCE)} 
Vocab.\ size: {[}256, 1024, 4096{]}; Sender entropy regularization coef., $\lambda_s$: {[}0.01, 0.025, 0.05, 0.1, 0.5, 1.0{]}; Receiver entropy regularization coef., $\lambda_r$: {[}0.01, 0.1, 0.5, 1.0{]}; learning rate: {[0.0001, 0.001, 0.01{]}; max.\ number of epochs: 1000; random seeds: {[}0, 1, 2, 3{]}; batch size: 2048; early stopping thr.: 0.99; bits shown to Receiver: {[}0, 1, 2, 3, 4, 5, 6, 7, 8{]}.

\textbf{Guess Number (Stochastic Computation Graph approach)}:
Vocab.\ size: {[}256, 1024, 4096{]}; Sender entropy regularization coef., $\lambda_s$: {[}0.01, 0.05, 0.1, 0.25{]}; learning rate: {[0.0001, 0.001{]}; max.\ number of epochs: 1000; random seeds: {[}0, 1, 2, 3{]}; batch size: 2048; early stopping thr.: 0.99; bits shown to Receiver: {[}0, 1, 2, 3, 4, 5, 6, 7, 8{]}.

\textbf{Image Classification experiments}
Vocab.\ size: {[}512,  1024, 2048{]}; temperature, $\tau$: {[}0.5, 0.75, 1.0, 1.5, 2.0{]}; learning rate: {[}0.001{]}, max.\ number of epochs: 100; random seeds: {[}0, 1, 2{]}; batch size: 32; early stopping thr.: 0.98; number of classes: {[}2, 4, 10, 20, 25, 50, 100{]}.

\textbf{Fitting random labels experiments}
Vocab.\ size: 1024; temperature, $\tau$: {[}1.0, 10.0{]}; learning rate: 0.0001, max.\ number of epochs: 200; random seeds: {[}0, 1, 2, 3, 4{]}; batch size: 32; early stopping thr.: $\infty$; prob.\ of label corruption: {[}0.0, 0.5, 1.0{]}.

\textbf{Adversarial attack experiments}
Vocab.\ size: 1024; temperature, $\tau$: {[}0.1, 1.0, 10.0{]}; learning rate: 0.0001, max.\ number of epochs: 200; random seeds: {[}0, 1, 2, 3, 4{]}; batch size: 32; early stopping thr.: 0.98.

\section{Evolution of message entropy during training}

In this Section, we aim to gain additional insight into development of the communication protocol by measuring its entropy during training. We concentrate on Guess Number and use the same 
experimental runs summarized in Figure~1 of the main text.

For each game configuration (that is, number of bits hidden from Receiver), we randomly select one successful run and plot the evolution of Sender message entropy and accuracy over training epochs.\footnote{We exclude the configuration in which Receiver sees the entire input, as it is a degenerate case of non-communication, as discussed in Section~4 of the main text.} We also plot entropy and accuracy curves for a randomly selected failed run, to verify to what extent entropy  development depends on task success.

We report results for runs where training was performed with Gumbel-Softmax relaxation and with the Stochastic Graph Computation approach in Figures~\ref{fig:gs_dynamics} and \ref{fig:rf_dynamics}, respectively. The reported entropy and accuracy values are calculated in evaluation mode, where Sender's output is selected greedily, without sampling. A higher entropy of such deterministic Sender indicates that the latter can encode more information about inputs in its messages. 

From these results, we firstly observe that the initial entropy of Sender's messages (before training) can be both higher than required for communication success (Figures~\ref{fig:gs_dynamics_6} and  \ref{fig:rf_dynamics_6}) and lower (the rest). When it starts higher than needed, it generally falls closer to the minimum level required for the solution. When the initial value is low, it increases during training. The failed runs can have message entropy above (Figures~\ref{fig:gs_dynamics_6}, \ref{fig:gs_dynamics_4} \& \ref{fig:rf_dynamics_6}) and below (e.g.\ Figures~\ref{fig:gs_dynamics_2}, \ref{fig:gs_dynamics_0} \& \ref{fig:rf_dynamics_0}) successful runs, suggesting that there is no systematic relation between degree of entropy and task success.

The fact that the entropy can be reduced with no decrease in accuracy or even with accuracy growth (e.g.\ Figure~\ref{fig:gs_dynamics_6}, red line, epochs 5..30) indicates that the tendency to discover new messages (increasing entropy) is counter-balanced by the complexity of mutual coordination with Receiver when entropy is larger. In our interpretation, it is this interplay that serves as a source of the natural bottleneck.

Finally, while in some runs the entropy is effectively increased w.r.t.\ its initialization level, the resulting protocol's entropy is at, or slightly above the lower bound of what the task allows. In this sense, we argue that the reported effect can be correctly denoted as a ``minimization'' result.

\begin{figure*}
\centering
\begin{subfigure}{1.0\linewidth}
\centering
  \includegraphics[width=0.8\linewidth]{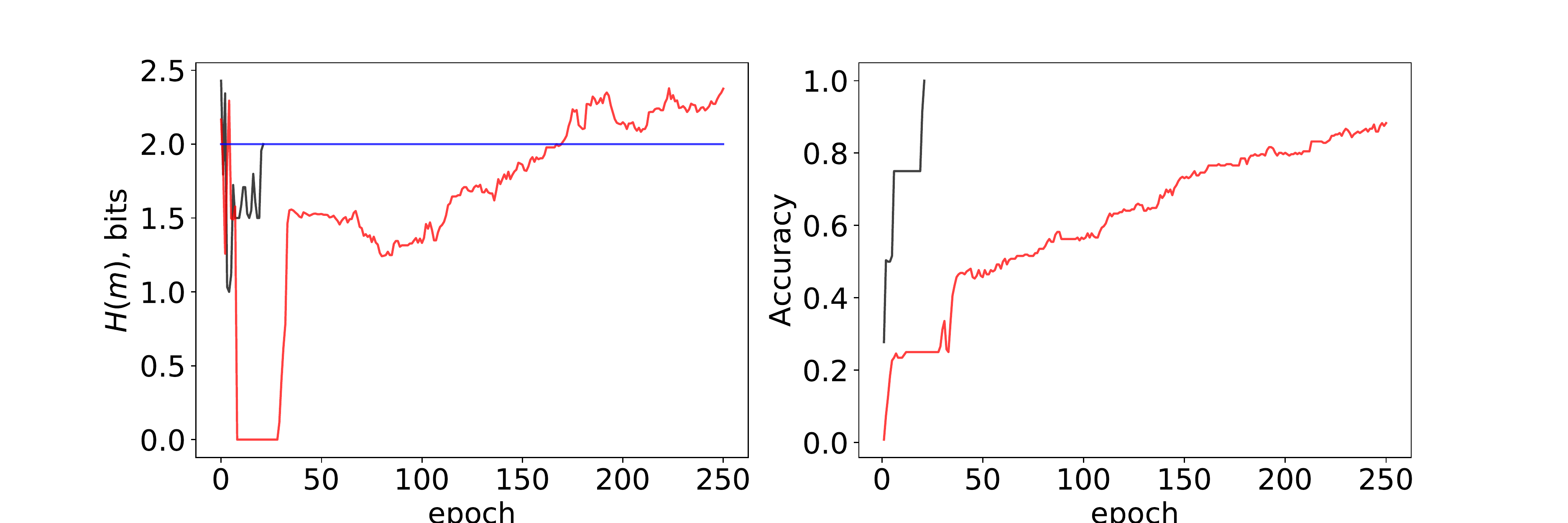}
    \caption{Binary digits hidden: 2}
  \label{fig:gs_dynamics_6}
 \end{subfigure}
 
\begin{subfigure}{1.0\linewidth}
\centering
  \includegraphics[width=0.8\linewidth]{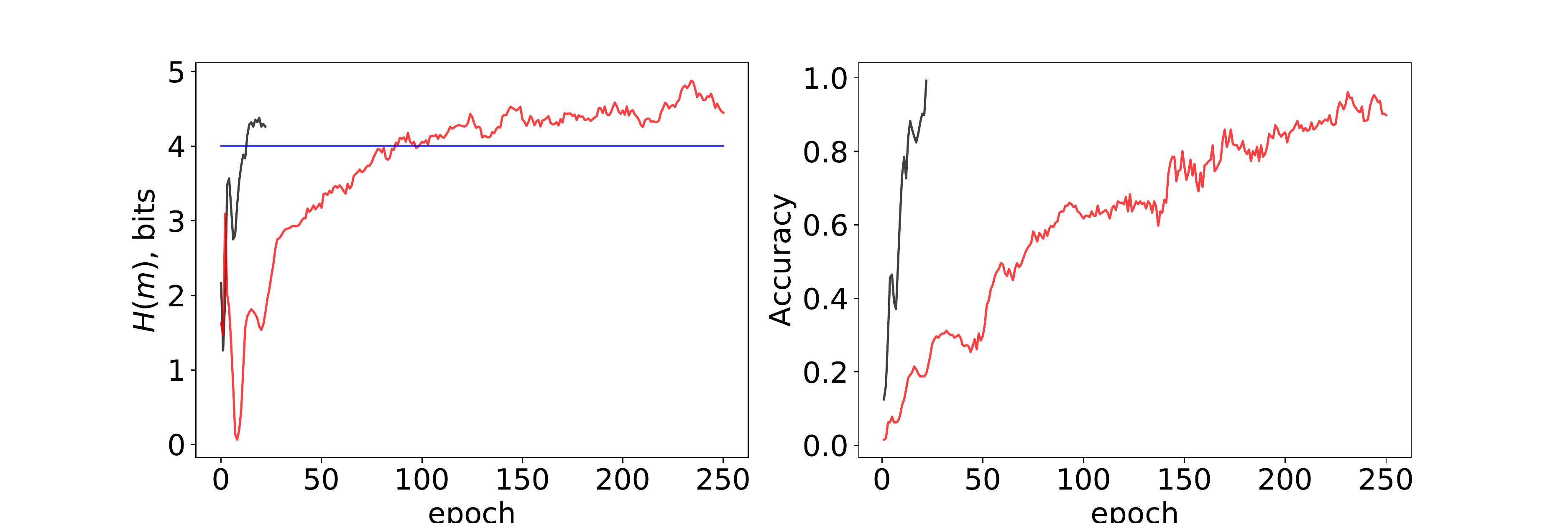}
    \caption{Binary digits hidden: 4}
  \label{fig:gs_dynamics_4}
\end{subfigure}
 
\begin{subfigure}{1.0\linewidth}
\centering
  \includegraphics[width=0.8\linewidth]{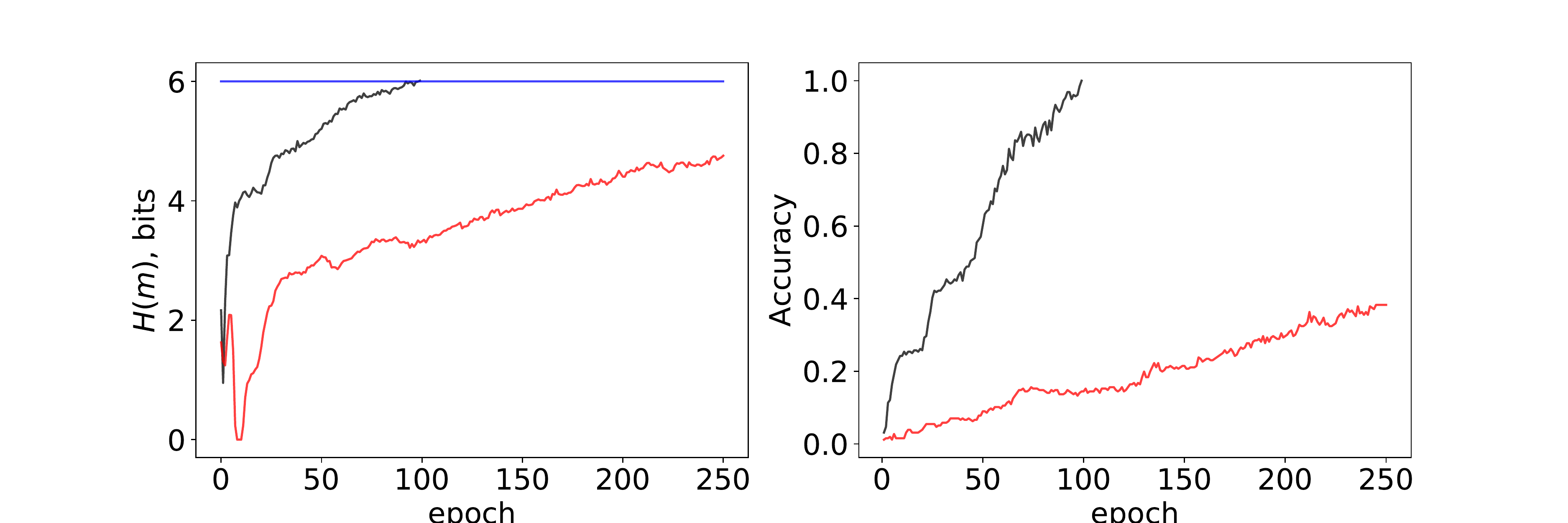}
    \caption{Binary digits hidden: 6}
  \label{fig:gs_dynamics_2}
 \end{subfigure}
 
 \begin{subfigure}{1.0\linewidth}
\centering
  \includegraphics[width=0.8\linewidth]{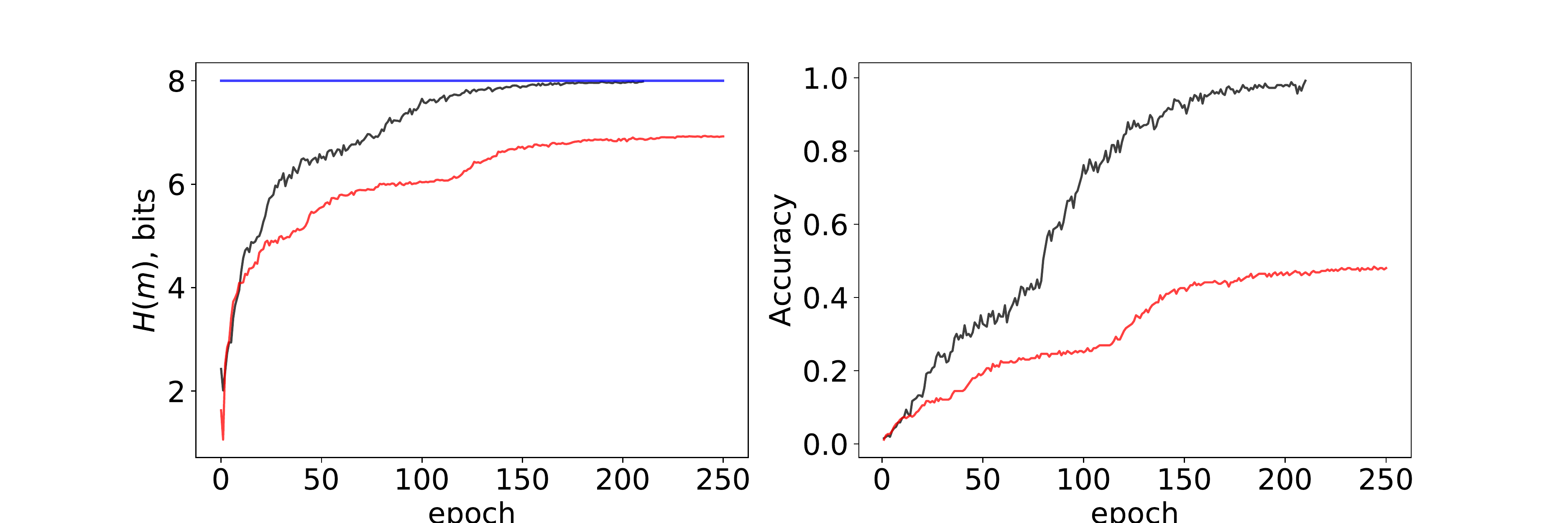}
    \caption{Binary digits hidden: 8}
       \label{fig:gs_dynamics_0}

 \end{subfigure}

\caption{Evolution of $H(m)$ over training epochs. Gumbel Softmax-based optimization, Guess Number. For each game configuration, specified by the number of bits Receiver lacks, we sample one successful (black line) and one failed (red line) training trajectory. The blue line marks $H_{min}$, minimal entropy for a successful solution.}
\label{fig:gs_dynamics}
\end{figure*}

\begin{figure*}
\centering
\begin{subfigure}{1.0\linewidth}
\centering
  \includegraphics[width=0.8\linewidth]{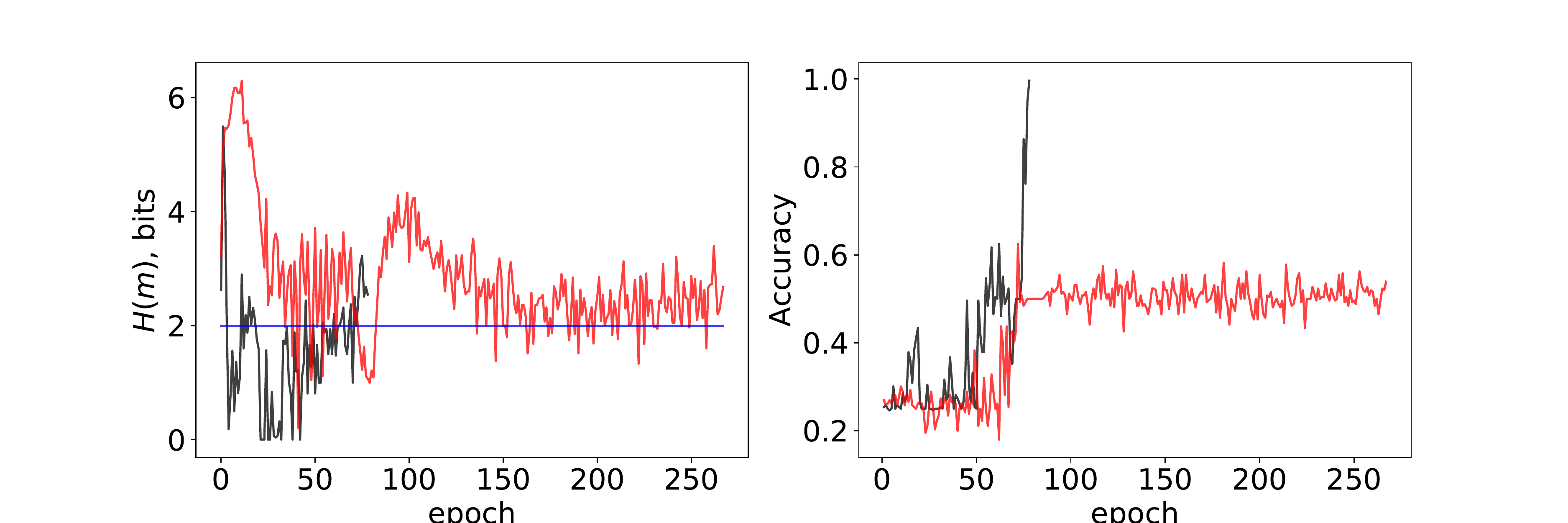}
  \caption{Binary digits hidden: 2}
    \label{fig:rf_dynamics_6}
 \end{subfigure}
 
\begin{subfigure}{1.0\linewidth}
\centering
  \includegraphics[width=0.8\linewidth]{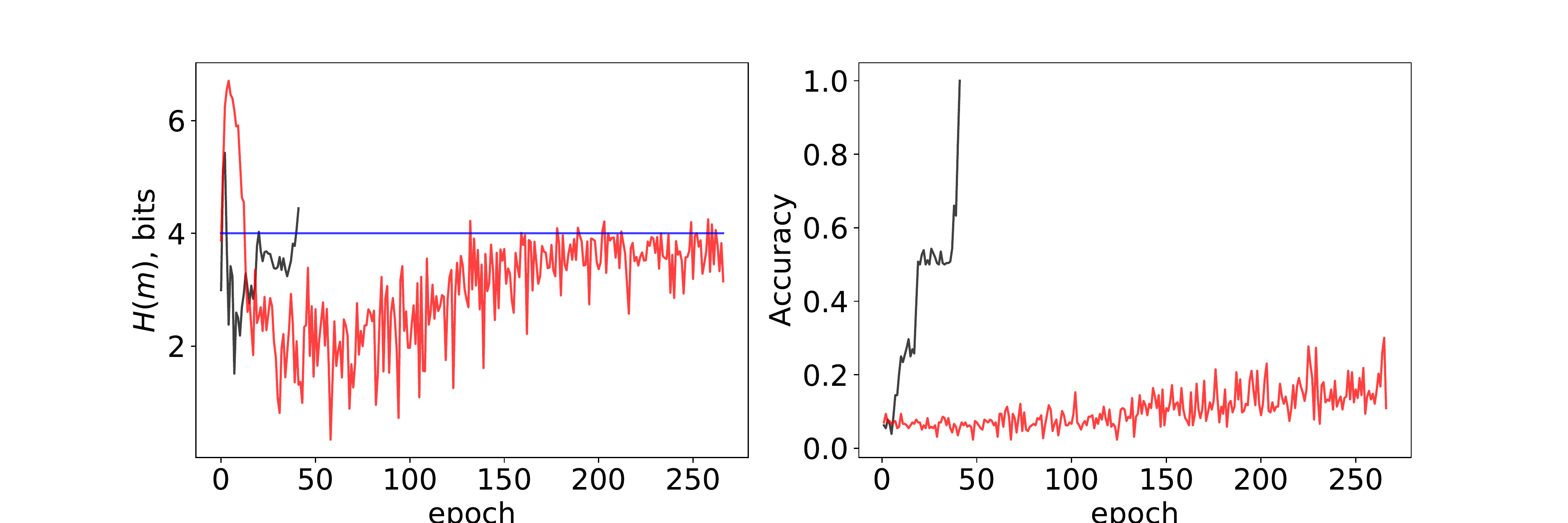}
    \caption{Binary digits hidden: 4}
        \label{fig:rf_dynamics_4}
 \end{subfigure}
 
\begin{subfigure}{1.0\linewidth}
\centering
  \includegraphics[width=0.8\linewidth]{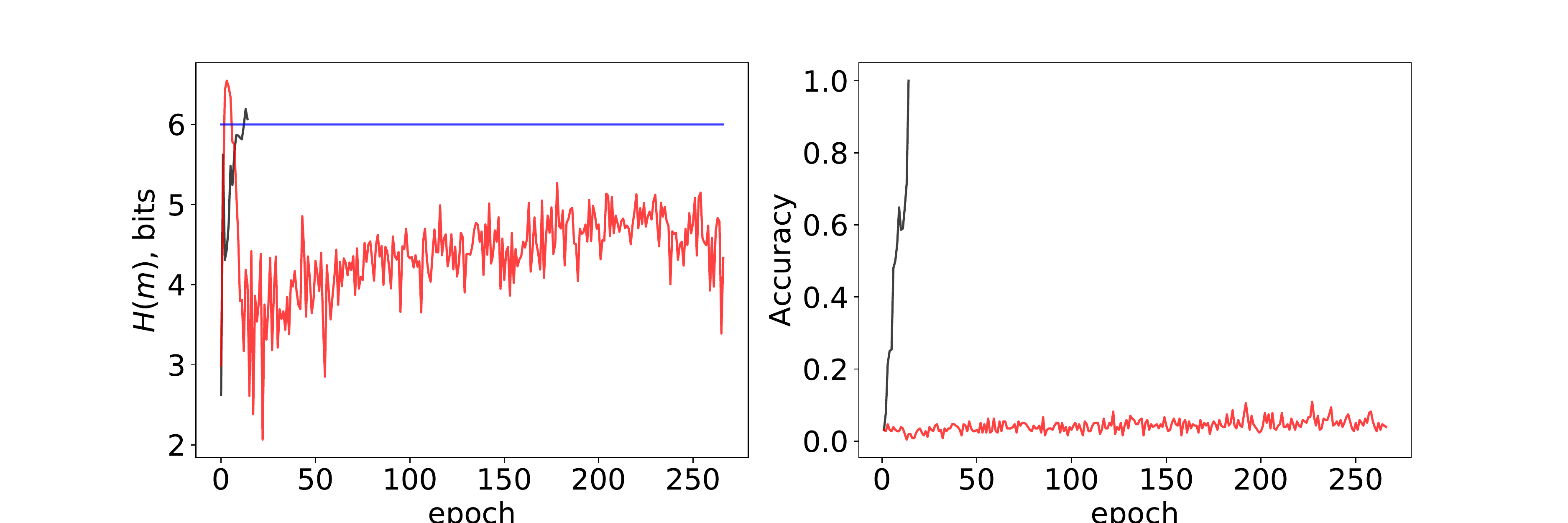}
    \caption{Binary digits hidden: 6}
 \end{subfigure}
 
 \begin{subfigure}{1.0\linewidth}
\centering
  \includegraphics[width=0.8\linewidth]{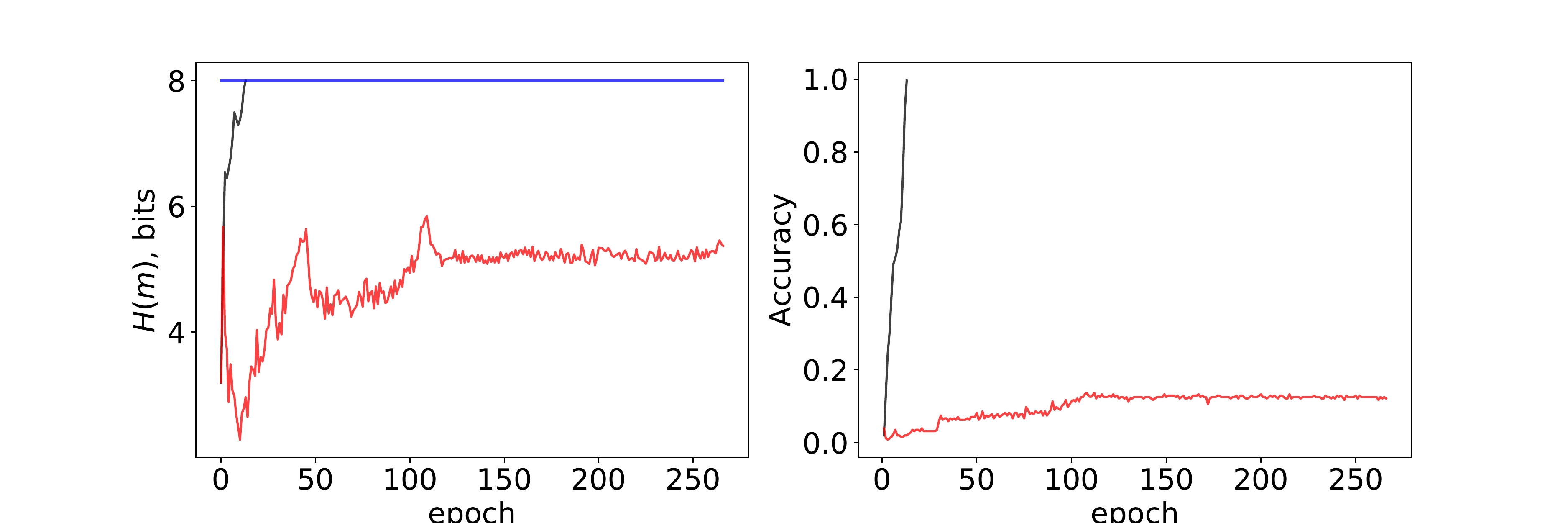}
    \caption{Binary digits hidden: 8}
        \label{fig:rf_dynamics_0}
 \end{subfigure}
\caption{Evolution of $H(m)$ over training epochs. Stochastic Computation Graph-based optimization, Guess Number. For each game configuration, specified by the number of bits Receiver lacks, we sample one successful (black line) and one failed (red line) training trajectory. The blue line marks $H_{min}$, minimal entropy for a successful solution.}
\label{fig:rf_dynamics}
\end{figure*}

\bibliographystyle{unsrtnat}
\bibliography{biblio,other}

\end{document}